\documentclass[letterpaper, 10 pt, journal, twoside]{IEEEtran}  %
\usepackage{amsmath,amsfonts,amssymb}
\let\labelindent\relax
\usepackage{enumitem}
\usepackage{prettyref}
\usepackage{graphicx}
\usepackage{wrapfig}
\usepackage[caption=false,font=footnotesize]{subfig}
\usepackage{array,multirow}
\usepackage{booktabs}
\usepackage[table]{xcolor}
\usepackage{cellspace}
\usepackage[T1]{fontenc}
\usepackage{comment}
\usepackage{url}
\usepackage{makecell}
\usepackage{hhline}
\usepackage
[backend=bibtex,
bibstyle=ieee,
citestyle=numeric,
sortcites,
natbib=true,
doi=false,
isbn=false,
url=true,
hyperref=true,
sorting=none,
eprint=false,
maxbibnames=99]{biblatex}

\usepackage{hyperref} 

\usepackage[ruled,vlined,noend,linesnumbered]{algorithm2e}

\newrefformat{Fig}{Fig.~\ref{#1}}
\newrefformat{fig}{Fig.~\ref{#1}}
\newrefformat{par}{Section~\ref{#1}}
\newrefformat{appen}{Appendix~\ref{#1}}
\newrefformat{sec}{Section~\ref{#1}}
\newrefformat{sub}{Section~\ref{#1}}
\newrefformat{table}{Table~\ref{#1}}
\newrefformat{alg}{Algorithm~\ref{#1}}
\newrefformat{Alg}{Algorithm~\ref{#1}}
\newrefformat{Def}{Definition~\ref{#1}}
\newrefformat{Thm}{Theorem~\ref{#1}}
\newrefformat{Lem}{Lemma~\ref{#1}}
\newrefformat{step}{Step~\ref{#1}}
\newrefformat{ln}{Line~\ref{#1}}
\newrefformat{eq}{Eqn.~\ref{#1}}
\newrefformat{eqn}{Eqn.~\ref{#1}}
\newrefformat{pb}{Problem~\ref{#1}}
\newrefformat{it}{Item~\ref{#1}}
\newrefformat{te}{Term~\ref{#1}}
\def\Eqref Eq:#1:{\eqref{eq:#1}}
\newrefformat{Eq}{Equation~\Eqref#1:}
\newcommand{\revise}[1]{\textcolor{black}{#1}}
\begin{document}
\title{Automated Heart and Lung Auscultation in Robotic Physical Examinations}
\author{Yifan Zhu$^{1}$, Alexander Smith$^{2}$, and Kris Hauser$^{1}$
\thanks{Manuscript received: September, 9, 2021; Revised December, 13, 2021; Accepted January, 11, 2022.}
\thanks{This paper was recommended for publication by Jessica Burgner-Kahrs upon evaluation of the Associate Editor and Reviewers' comments. This work was supported by NSF Grant \#2025782.}
\thanks{$^{1}$: Y. Zhu and K. Hauser are with the Departments of Computer Science, University of Illinois at Urbana-Champaign, IL, USA.
        {\tt\small \{yifan16, kkhauser\}@illinois.edu}}%
\thanks{$^{2}$: A. Smith is with the Carle Illinois College of Medicine, IL, USA.
        {\tt\small ads10@illnois.edu}}%
\thanks{Digital Object Identifier (DOI): see top of this page.}
}

\markboth{IEEE Robotics and Automation Letters. Preprint Version. Accepted January, 2022}
{Zhu \MakeLowercase{\textit{et al.}}: Automated Heart and Lung Auscultation in Robotic
Physical Examinations} 
\maketitle

\begin{abstract}
This paper presents the first implementation of autonomous robotic auscultation of heart and lung sounds.  To select auscultation locations that generate high-quality sounds, a Bayesian Optimization (BO) formulation leverages visual anatomical cues to predict where high-quality sounds might be located, while using auditory feedback to adapt to patient-specific anatomical qualities. Sound quality is estimated online using machine learning models trained on a database of heart and lung stethoscope recordings. Experiments on 4 human subjects show that our system autonomously captures heart and lung sounds of similar quality compared to tele-operation by \revise{a human trained in clinical auscultation. Surprisingly, one of the subjects exhibited a previously unknown cardiac pathology that was first identified using our robot, which demonstrates the potential utility of autonomous robotic auscultation for health screening.}
\end{abstract}


\section{Introduction}
\IEEEPARstart{I}{nfectious} diseases pose substantial risks to healthcare providers \revise{and there have been worldwide initiatives to use robots and automation to help combat the COVID-19 pandemic~\cite{Khamis2021}. In particular,} performing physical examinations with robots has potential to reducing the risks to healthcare providers by minimizing person-to-person contact. Moreover, robotic health screening can also help promote preventative care and affordable routine checkups, particularly in rural, remote, and low-resource communities.

\begin{figure}[tbp]
\centering
\includegraphics[width=.96\linewidth]{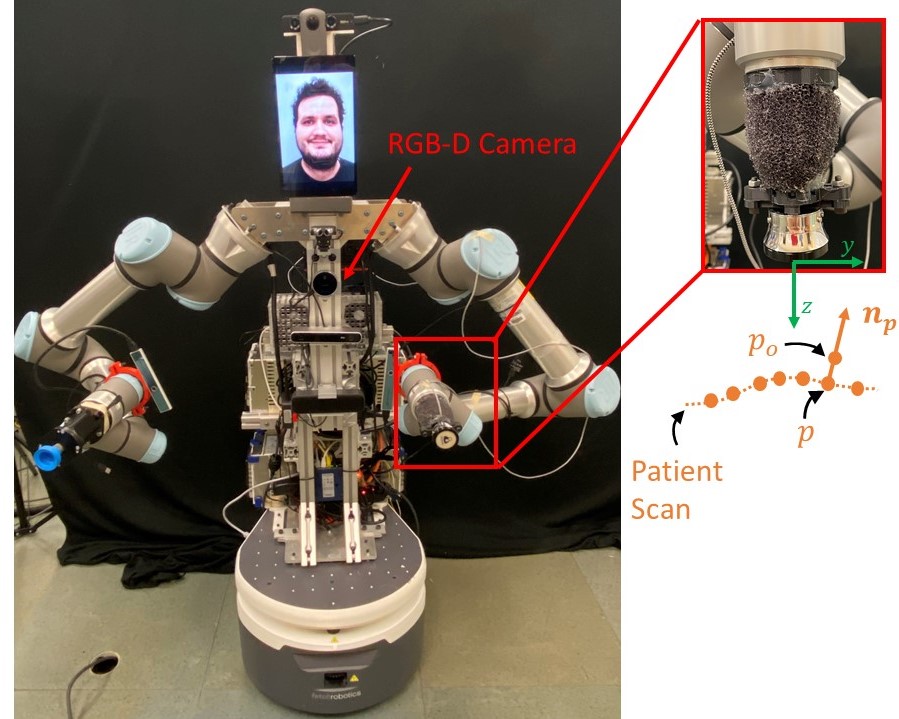}
\caption{\label{fig:TRINA} The TRINA robot used in the experiments, with the zoomed-in view of the stethoscope on the right. The stethoscope frame is shown in green, and the patient scan point cloud in orange. [Best viewed in color.]}
\vspace{-10px}
\end{figure}

\revise{Motivated by the benefits of automating physical examinations,} this paper explores the task of auscultation, i.e., listening to heart and lung sounds via a stethoscope, which which is an important screening procedure that is performed at regular checkups and for patients exhibiting respiratory and cardiac symptoms. This paper demonstrates the first robotic system capable of performing automated heart and lung auscultation, based on the TRINA robot shown in~\prettyref{fig:TRINA}. Prior tele-medicine robotic systems have been built for performing auscultation and echocardiography~\cite{Giuliani2020,Yang2020} through tele-operation. Compared to  tele-operation, automated medical exams offer several potential benefits including lower operation time, a shorter learning curve for physicians, and reduced cognitive load and tedium. 
Auscultation is, however, a challenging process to automate. The quality and interpretability of sounds produced by the anatomical structure of interest at the skin surface is highly variable and depends on the location, pressure, and steadiness of the stethoscope. If the stethoscope is placed at a poor location in relation to the patient's internal anatomical structures (e.g., over a rib or deep layers of fatty tissue), the sound may be attenuated or muffled in frequencies that are critical for diagnosis. A human doctor uses visual, tactile, and audio feedback as well as anatomical information and diagnostic expertise from prior medical training to localize informative listening locations.

The goal of our system is to record high-quality sounds that a human doctor will listen to, so the robot's aim is to provide sounds of diagnostic utility rather than to perform diagnosis itself. First, our system captures a 3D point cloud scan of the patient, registers a human body model, estimates the locations of key anatomical landmarks, and produces a prior map of high-quality auscultation locations. It then adopts informative path planning using audial feedback to adaptively search over the region of interest for a high-quality auscultation location. Audial feedback relies on {\em sound quality estimators} trained on a database of heart and lung stethoscope recordings. To determine the optimal sensing location we formulate a Bayesian Optimization~(BO) problem where the unknown sound quality field is estimated as a semi-parametric residual Gaussian Process (SPAR-GP)  model, with a prior map that depends on latent translation offset and sound quality scaling parameters.

Experiments on 4 healthy male human subjects demonstrate that our system performs heart and lung auscultation automatically, and locates sounds of similar diagnostic quality and execution time as tele-operation by a  \revise{human trained in auscultation in medical school, and trained to use the robot by the investigators.} The procedure is reliable and largely required no supervision, except for one subject where a perception error required anatomical landmarks to be input manually before ausculation. Although each subject reported good health and no respiratory or cardiac ailments, in a surprising discovery one subject exhibited a heart murmur that was identified during these experiments. The subject was thereafter referred to a physician for follow-up tests.

\section{Related Work}
\subsection{Robotic Remote Auscultation and Sonography}
Tele-nursing robots consisting of a mobile base and arms have  demonstrated a variety of nursing tasks including handovers, vital signs monitoring, disinfection, and auscultation~\cite{LiTRINASystem2017, arent2017selected, Kim2021, Yang2020}. Robotic technologies employed in the fight against COVID-19 and infectious diseases are reviewed in recent surveys~\cite{Wang2021,DiLallo2021}. 

Giuliani et al. develop a robot and tele-operation user interface for echocardiography~\cite{Giuliani2020}. Several works have studied performing robot remote lung sonography using the MGIUS-R3 robotic tele-echography system produced by MGI Tech Co, Ltd.~\cite{Evans2020,Adams2020,Yu2020,Wang2021b}. Marthur et al. develop a semi-autonomous robotic system to perform trauma assessment remotely, where the locations on a patient to be assessed are identified automatically via perception and sonography is performed with tele-operation~\cite{Mathur2019}. Instead of tele-operation, our proposed method performs auscultation fully automatically, which \revise{has the potential to} reduce physician learning curve, cognitive load, and tedium. \revise{On a level of autonomy (LoA) scale from 0 (Full Manual) to 5 (Full Autonomy)~\cite{Haidegger2019}, our system achieves a  level of 4 (High-level Autonomy) because the robot autonomously performs the procedure while an operator monitors and intervenes when necessary.}

\subsection{Informative Path Planning}
Our problem falls under the category of informative path planning (IPP), where the robot plans information-gathering paths given a probabilistic model of the quantity of interest. Sensor placement, active learning, and adaptive sampling problems can be seen as special instances of the IPP problem.  IPP has been applied to a variety of robotics applications such as object tactile exploration~\cite{Driess2019}, palpation-based tissue abnormalities detection~\cite{Salman2018,Ayvali2017,Goldman2013}, inspection and environment mapping~\cite{Hollinger2013,Hollinger2013a,Binney2012}.

Most related to ours are the works by Salman et al.~\cite{Salman2018} and Ayvali et al.~\cite{Ayvali2017}. Salman et al. propose both discrete and continuous trajectory search algorithms for robots to search for the boundaries of tissue abnormalities, i.e, high stiffness, through robotic palpation. Gaussian process (GP) regression is used to model the distribution of stiffness. Four methods, including BO, active area search, active level sets estimation, and uncertainty sampling are compared, among which active area search had the best performance. Ayvali et al. also perform robot palpation to detect tissue abnormalities, using GP regression and BO. To inject user prior knowledge of the locations of tissue abnormalities, they add a decaying utility function whose value peaks at the user-provided locations to the acquisition function of BO. In contrast, our work adopts a discrete search BO to the auscultation setting, and introduces two technical contributions to make auscultation more practical. First, visual registration of anatomical landmarks eliminates the need for manual input to specify priors, except as a backup for registration failures, and second, the use of our SPAR-GP model leads to faster convergence than pure residual GP models. 

\revise{Belonging to the family of semi-parametric regression methods, SPAR-GP performs function approximation with a combination of parametric functions and Gaussian process. In robotics, similar models have been applied to  system identification of linear and nonlinear dynamical systems~\cite{Wu2012,Ko2007,Camoriano2016}, but to our knowledge semi-parametric models have not been adopted in the informative path planning setting.}

\section{Method}

\begin{figure}[tbp]
\centering
\includegraphics[trim=0cm 0cm 0cm 0cm,clip,width=0.99\linewidth]{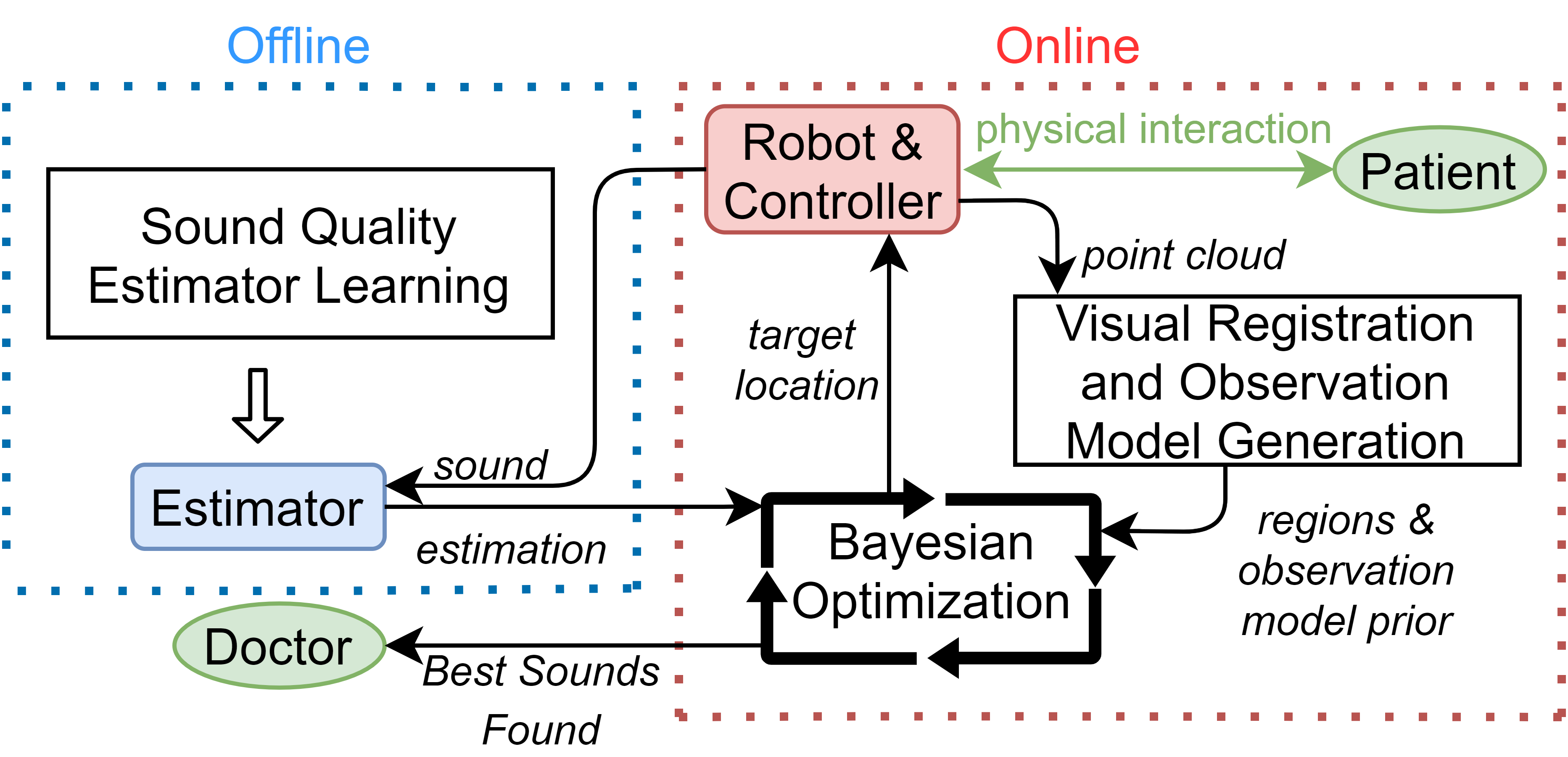}
\caption{\label{fig:flowchart} The workflow of our method.}
\vspace{-3mm}
\end{figure}

The overview of our method is shown in~\prettyref{fig:flowchart}. In the \textit{offline phase}, we build a dataset of heart and lung stethoscope recordings of humans with labeled sound qualities and train heart and lung sound quality estimators (\prettyref{sec:Estimator}). The goal of our system is to generate high (estimated) quality heart and/or lung recordings by placing a stethoscope at anatomically relevant portions of a patient's chest and back, while keeping the number of auscultation actions small. These estimators are the only feedback used by the system during the online exam; we do not assume human raters are available to provide ground truth.

During the \textit{online phase}, the robot first performs visual registration of the patient to a reference human model with labeled \textit{clinical auscultation locations}. An \textit{observation model prior} is constructed based on the visual registration (\prettyref{sec:registration}), which encodes anatomical prior information about expected sound quality. Finally we use BO (\prettyref{sec:formulation}) to select locations to auscultate to find the best sounds utilizing both the learned sound quality estimators and prior information from visual cues. These targets are then used for motion control (\prettyref{sec:motion}).

\subsection{BO Formulation}\label{sec:formulation}
We use BO to search adaptively for an auscultation location that yields a high quality sound within a specified anatomical region. We denote sound quality estimators for heart and lung sounds as $e_s(r)$, where $s$ is an anatomical structure (heart or lung) and $r$ is a stethoscope recording.  The BO approach computes a probabilistic estimate of the unknown field $e_s(r(x))$ across the patient surface using a SPAR-GP model, which is a sum of the observation prior and the GP. An acquisition function is optimized to yield the new auscultation location. Given a new observation, the estimate is re-fit to the data and the process repeats until a termination criterion is met. 

\subsubsection{SPAR-GP}
A GP models an unknown function $f$ as a collection of random variables $f(x)$ which are jointly Gaussian when evaluated at locations $x$. A GP is fully specified by its mean function $m(\cdot)$, which we set to $0$, and covariance function $k(\cdot,\cdot)$:
\begin{equation*}
    f(x) \sim \mathcal{GP}(m(x),k(x,x'))
\end{equation*}

Given $n$ existing observed function values $\Bar{\mathbf{y}} = [y_1,\dots,y_n]$ at $\Bar{\mathbf{x}} = [x_1,\dots,x_n]$, GP regression predicts the function values at new point $x^*$ as a Gaussian distribution:
\begin{equation*}
\begin{split}
    P(f(x^*)|\Bar{\mathbf{x}}, \Bar{\mathbf{y}}, x^*) \sim \mathcal{N}(\mathbf{k}\mathbf{K}^{-1}\Bar{\mathbf{y}},k(x^*,x^*) - \mathbf{k}\mathbf{K}^{-1}\mathbf{k}^T )\\
\end{split}
\end{equation*}
Here,
\begin{equation*}
    \mathbf{K} = \begin{bmatrix}
                k(x_1,x_1) & \cdots & k(x_1,x_n)\\
                \vdots & \ddots & \vdots\\
                k(x_n,x_1) & \cdots & k(x_n,k_n)\\
                \end{bmatrix} + \sigma^2\mathbf{I}
\end{equation*}
\begin{equation*}
    \mathbf{k} = \begin{bmatrix}
                k(x^*,x_1), & \cdots, & k(x^*,x_n)\\
                \end{bmatrix},
\end{equation*}
where $\sigma$ is the standard deviation of noise at an observation. 

For each anatomical location, the observation model is a sum of a parametric prior mean function $\mu_{\theta}(x)$ and a GP residual function $f_s(x)$: $e_s(r(x)) \approx \mu_{\theta}(x) + f_s(x)$. Since the GP models residuals with respect to the prior, we subtract the prior from the sound quality as the GP observations: $y_i = e_s(r(x_i))-\mu_{\theta}(x_i)$. Here we further denote the history of the estimated sound qualities as $\Bar{\mathbf{e}} = [e_1,\dots,e_n]$.

Although the anatomical reference provides sound quality peaks assuming an average human and perfect registration of anatomical landmarks, the robot's prior should capture the uncertainty in visual registration error and the effect a patient's body type has on the magnitude of the overall sound quality. Therefore, we make the prior $\mu_\theta$ a parametric function of the latent variables $\theta$ representing the translation offset and sound quality scaling, and infer $\theta$ from observed sound qualities. In particular, a reference quality map $\mu_{o}(x)$ is first generated from visual registration and $\theta$ is initialized to $\theta_o$, which includes zero translation offset and scaling of 1. The exact composition of the prior mean function is deferred to~\prettyref{sec:registration}. $\theta$ is inferred after each reading using the history $\Bar{\mathbf{x}}$ and $\Bar{\mathbf{e}}$ with maximum a posteriori (MAP) estimation. We use a likelihood function  $\mathcal{L}(\theta|\Bar{\mathbf{x}},\Bar{\mathbf{e}}) = \prod g(e_i|\mu_{\theta}(x_i),\sigma^2)$, where $g(\cdot|\mu_{\theta}(x_i),\sigma^2)$ is the probability  density function of the Gaussian distribution $\mathcal{N}(\mu_{\theta}(x_i),\sigma^2)$. The prior of $\theta$, $h(\theta)$, follows a multivariate Gaussian distribution $\mathcal{N}(\theta_o,\Sigma)$. We solve the MAP estimation problem by maximizing the posterior,  using a standard numerical optimization solver:
\begin{equation}
\begin{split}
    \theta^* = \underset{\theta}{\text{argmax}} \, \mathcal{L}(\theta|\Bar{\mathbf{x}},\Bar{\mathbf{e}})h(\theta) \\
\end{split}
\end{equation}

\subsubsection{BO}
Letting $A$ be a region of interest on the patient surface near structure $s$, BO aims to solve:
\begin{equation}\label{eq:BO_obj}
\begin{split}
    \underset{x\in A}{\text{max}} \, e_s(r(x)) \\
\end{split}
\end{equation}

Since heart and lung sounds need to be listened separately and there are usually multiple parts of the anatomical structures that are of interest, we solve~\prettyref{eq:BO_obj} for each anatomical structure $s$ separately, and for each $s$ sequentially for their regions of interest. We also share the same GP across the entire chest or back of a patient for the same $s$.

In each iteration of BO, an acquisition function is used to determine the next location to observe the data. Two popular choices are expected improvement (EI) and upper confidence bound (UCB). Let the posterior mean and variance of the GP be $\mu_{\bar{y}}(x),\sigma_{\bar{y}}^2(x)$, then EI is defined as:
\begin{equation}
    EI(x) = \mathbb{E}[\text{max}(0,\mu_{\theta}(x) + \mu_{\bar{y}}(x)-e^*)],
\label{eq:EI}
\end{equation}
where $e^*$ is the best observed quality so far, and EI can be calculated in closed form~\cite{Brochu2010}. UCB is defined as: 
\begin{equation}
    UCB(x) = \mu_{\theta}(x) + \mu_{\bar{y}}(x) + \beta \sigma_{\bar{y}}(x),
\label{eq:UCB}
\end{equation}
where $\beta$ is an exploration weight that regulates how much bonus is given to uncertainty in the prediction. 

The overall algorithm is outlined in \prettyref{alg:BO}. For each iteration in each region, a point with the largest acquisition function value is first chosen to be observed, where $\xi(x)$ represents the acquisition function, and the quality is estimated. In particular, $A$ contains sampled points on a patient surface, and the maximization is performed in a brute force fashion by calculating the acquisition function values across all points in $A$ and selecting the maximum. In addition, the same point is observed at most once. Then we add the estimated qualities to the history and update the prior mean function by solving the MAP estimation, after which the residuals for GP are calculated, observations are added, and the GP is re-fit. For the termination criteria, we allow early termination if the estimated quality in a region is above an adequacy threshold for making a diagnosis. We also set the budget $N_{max}$ to a reasonable number of auscultations per region to limit the maximum time of the procedure.

\begin{algorithm}[h]
\SetAlgoLined
\textbf{Input:} Structure $s$ (heart/lung), prior $\mu_{\theta}$, regions $A$, max iterations $N_{max}$\;
Initialize $\Bar{\mathbf{x}} = \{\}$, $\Bar{\mathbf{y}} = \{\}$, $\Bar{\mathbf{e}} = \{\}$\;
\For{all regions $A$ near $s$}{
  \For{$k = 1,...,N_{max}$ }{
    $x_k \leftarrow \arg\max_{x \in A} \xi(x)$ \;
    \eIf{region termination criteria met}{
        Go to next region\;
    }{
        Auscultate at~$x_k$, obtain sound quality $e_k$\;
        Set $\Bar{\mathbf{x}} \gets \Bar{\mathbf{x}} \cup \{x_k\}$, $\Bar{\mathbf{e}} \gets \Bar{\mathbf{e}} \cup \{e_k\}$\;
        $\theta \gets \text{argmax} \, \mathcal{L}(\theta|\Bar{\mathbf{x}},\Bar{\mathbf{e}})h(\theta)$\; 
        Set $\Bar{\mathbf{y}} \gets \Bar{\mathbf{y}} \cup \{ e_k - \mu_{\theta}(x) \}$\;
        Re-fit GP\;
    }
  }
}
 \Return  recordings of max quality in each region\;
 \caption{Bayesian Optimization Auscultation}
 \label{alg:BO}
\end{algorithm}

\subsection{Sound Quality Estimator}\label{sec:Estimator}
\subsubsection{Dataset}
We first collect lung and heart stethoscope recordings of various qualities with TRINA from the researchers on this project. Each recording lasts 10\,s, and we vary the locations on the subject's chest, whether or not the subject is wearing clothing, and whether the subject is taking a deep breath. 80 recordings are collected each for heart and lung sounds. \revise{The ground truth qualities of the recordings were labeled by 4 medical school students who have undergone auscultation training in a Liaison Committee on Medical Education (LCME)-accredited medical program and have been applying their auscultation skills in clinical settings for at least one year. We note that prior studies suggest that the cardiac examination skills of trained medical school students are no worse and may even be better than those of experienced doctors~\cite{vukanovic2006competency}. } Each label is a score between 0-1 with increments of 0.125 based on the rating guidelines provided by Grooby et al., where 0 means no detectable heart or lung signal, and 1 means clear heart or lung sounds with little to no noise~\cite{Grooby2020}. A recording of poor quality either contains weak signals or has noise that obscures the signals. The noise typically comes from the stethoscope contacting a patient and heavy breathing sounds obfuscating heart sounds.

We use intra-class correlation (ICC)~\cite{Koo2016} to measure the inter-rater reliability of the labels. We obtain the ICC estimates using a mean-rating, absolute-agreement, 2-way mixed-effects model, using the Pingouin package at~\url{https://pingouin-stats.org/}. The calculated ICC estimates are 0.925 and 0.679, respectively for heart and lung, which correspond to excellent and moderate agreement~\cite{Koo2016}. The average variances for the 4 ratings of each stethoscope recording's quality are 0.0163 and 0.0400 for heart and lung respectively. The max variances are 0.0938 and 0.151 for heart and lung. The average of the 4 ratings of each recording is used as the quality label.  We further augment the dataset with 5 synthetic recordings of pure noise with varying amplitudes and labeled with quality 0, which emulate erroneous conditions like placing a stethoscope on irrelevant parts of the body, rubbing the stethoscope, bumping the stethoscope, and failing to make contact.

\subsubsection{Feature Extraction}
We follow the method of Grooby et al. for heart and lung quality classification and modify it for regression. For each sound recording, we apply noise reduction~\cite{Sainburg2020}, band-pass filter with cutoff frequency 50-250\,Hz for heart recordings and 200-1000\,Hz for lung recordings, and extract for all sound recordings the top features listed by Grooby et al. In particular, we use the top 5 features for heart sound and the top 2-6 features for lung sound (top 1 feature for lung is not used due to the lack of open-source code). The extracted features are then subsequently used for training the estimators.

\subsubsection{Learning}
\begin{figure}[tbp]
    \vspace{-7px}
    \centering
    \subfloat[\centering Heart]{{\includegraphics[trim=0cm 0cm 0cm 3.1cm,clip,width=4.4cm]{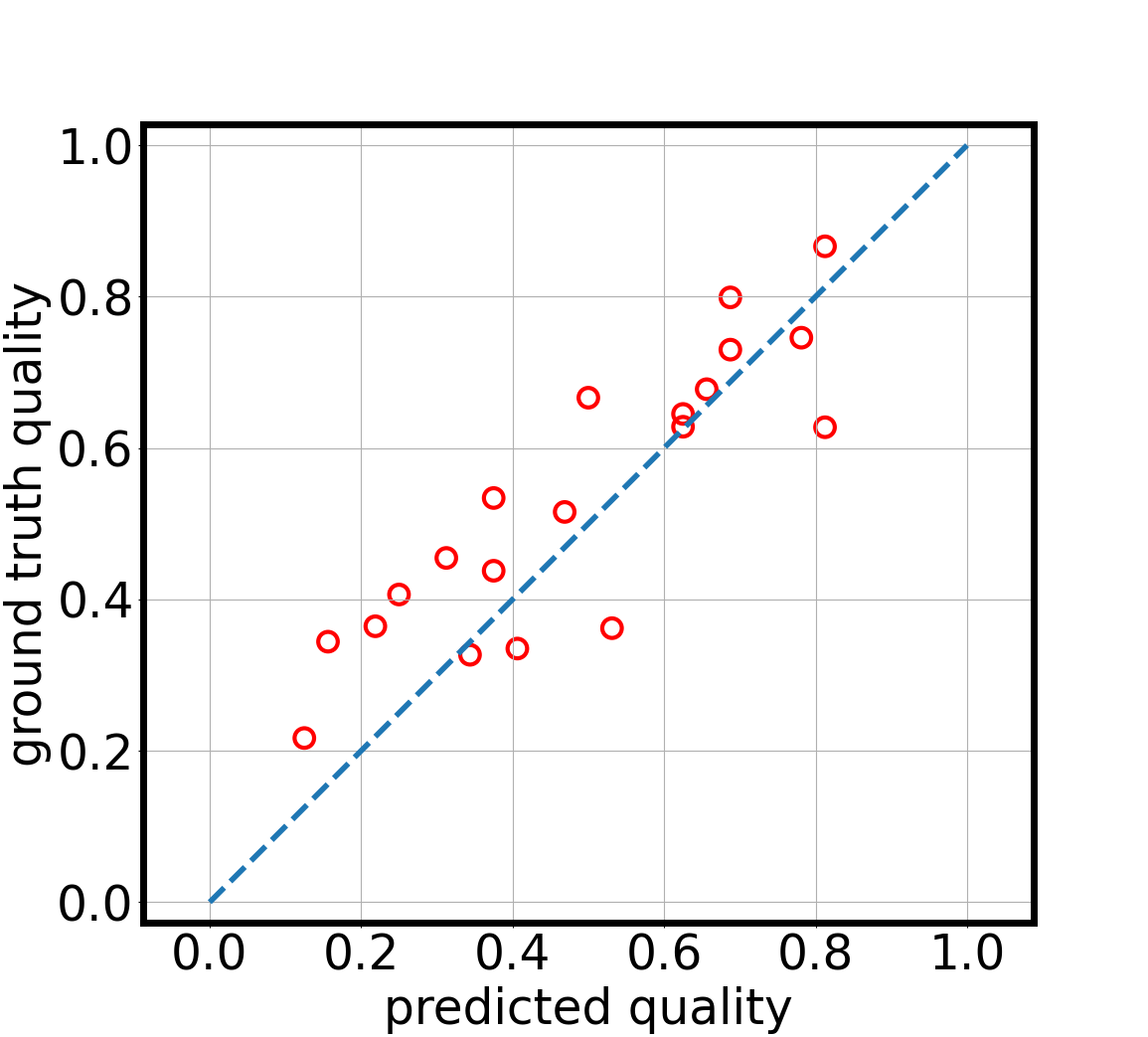} }}
    \subfloat[\centering Lung]{{\includegraphics[trim=0cm 0cm 0cm 3.1cm,clip,width=4.35cm]{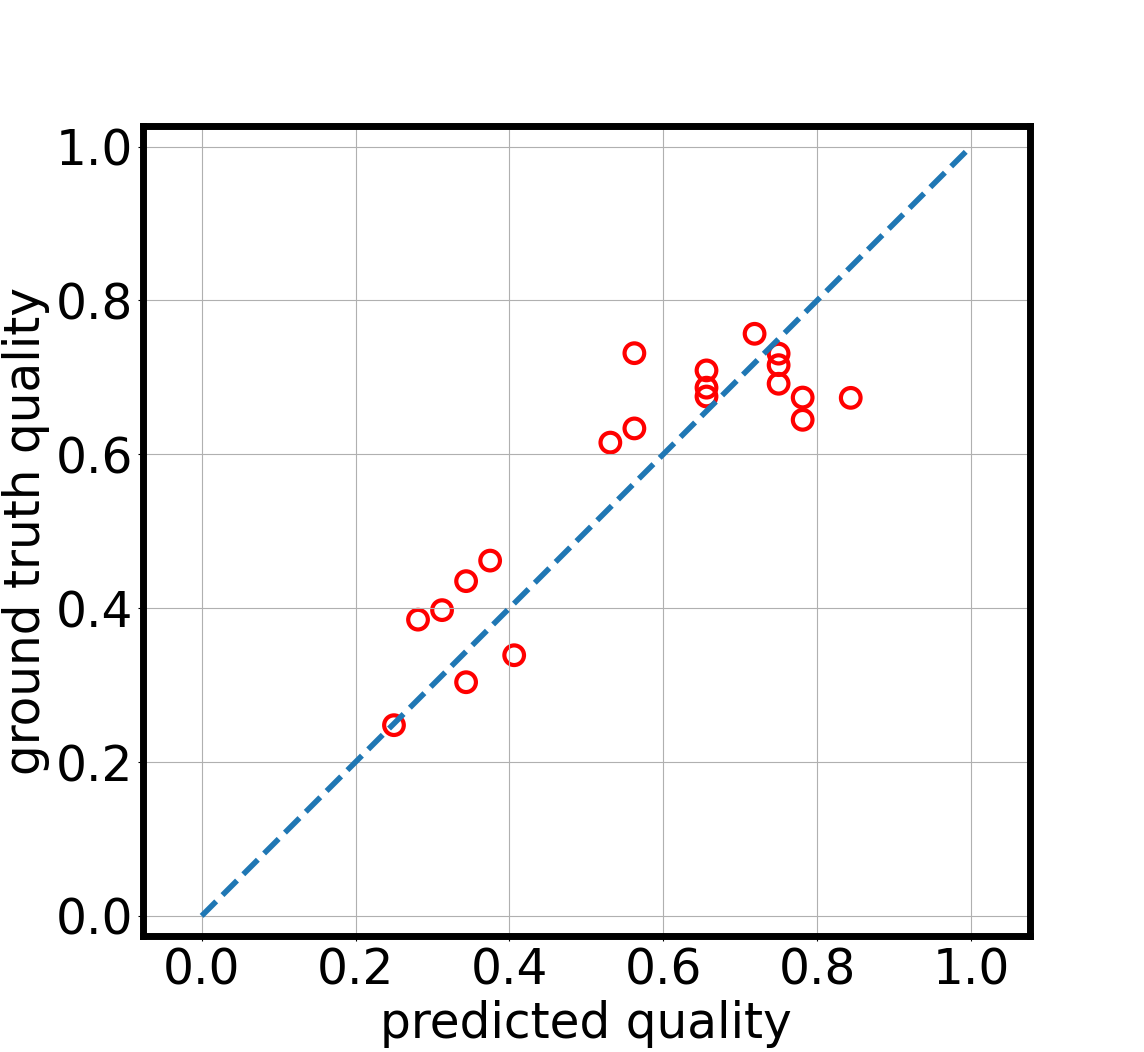} }}
    \caption{Trained heart and lung sound quality estimator predictions on the testing set, where the horizontal axis is the predictions and the vertical axis the ground truth labels.}
    \label{fig:predictions}
    \vspace{-5px}
\end{figure}

We use the TPOT AutoML algorithm~\cite{Le2020} to train the heart and lung sound quality estimators that take in the extracted features and predict the sound qualities. We split the dataset and use 25\% as testing data. For the TPOT algorithm, we set generations = 100, population size = 100, and set the rest of the hyperparameters as default. We achieve a 0.0945 and 0.0733 mean absolute errors~(MAE) on the testing set of heart and lung sounds respectively. The predictions on the testing set are shown in~\prettyref{fig:predictions}.

\subsection{Visual Registration and Sound Quality Prior}\label{sec:registration}
\begin{figure}[]
\centering
\setlength{\tabcolsep}{0px}
\begin{tabular}{ccc}
{\includegraphics[width=.3\linewidth]{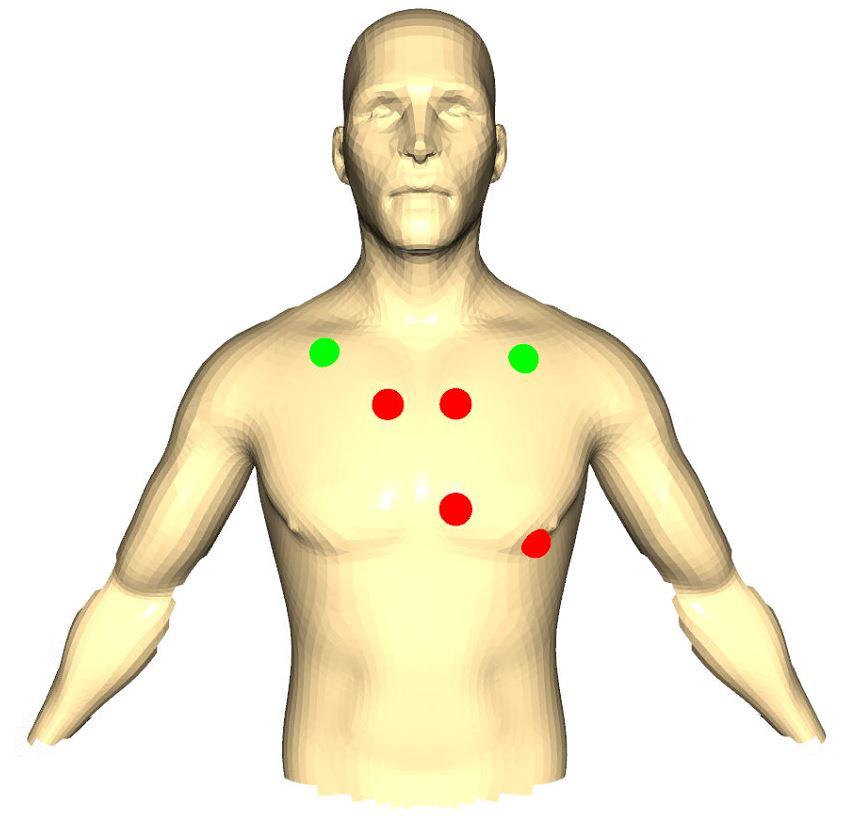}}\put(-65,74){\scalebox{0.8}{Reference Model}} &
{\includegraphics[width=.27\linewidth]{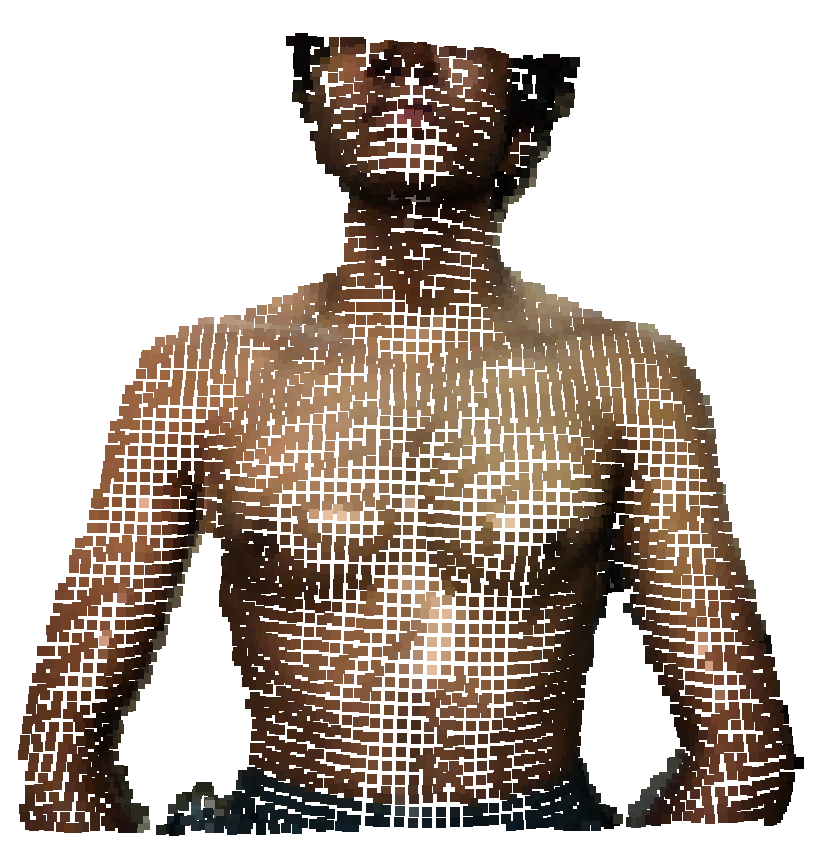}}\put(-58,74){\scalebox{0.8}{Patient Scan}} &
{\includegraphics[width=.335\linewidth]{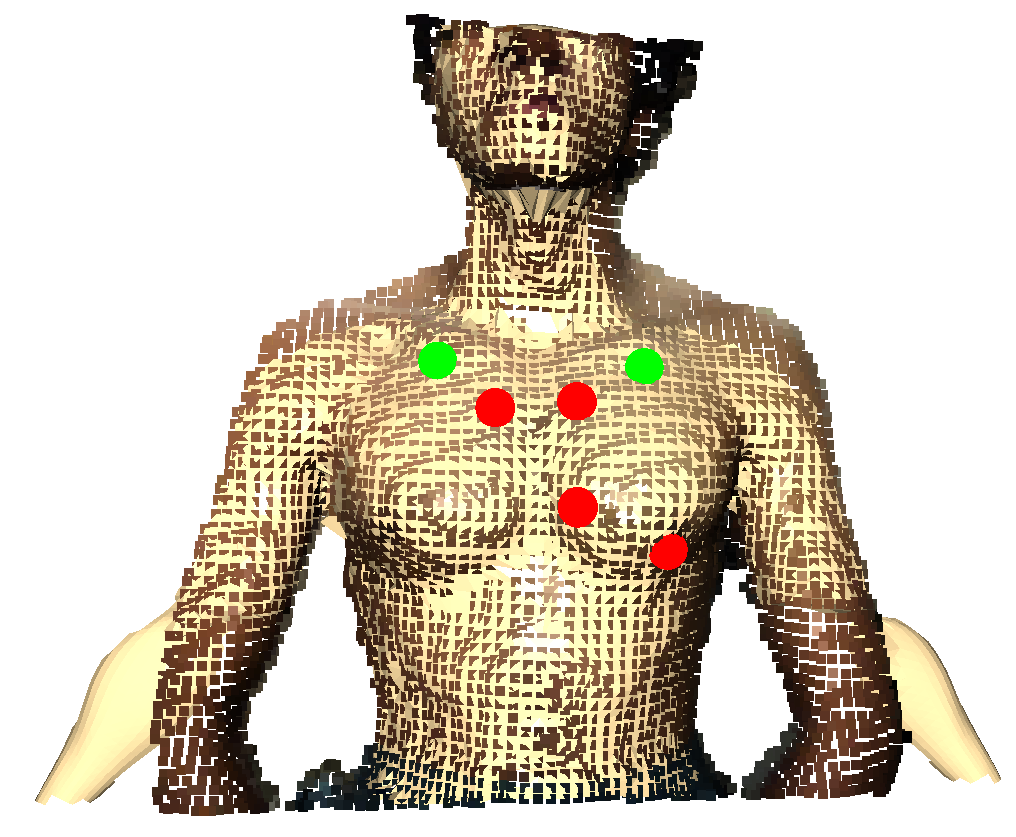}}\put(-70,74){\scalebox{0.8}{Registration Result}}   \\
{\includegraphics[width=.34\linewidth]{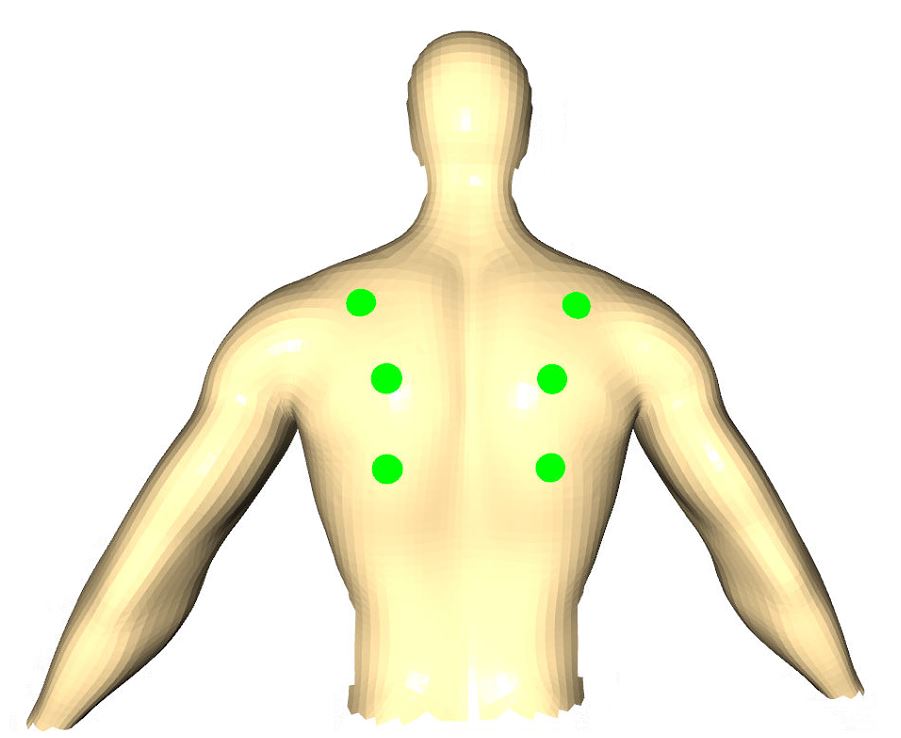}} &
{\includegraphics[width=.27\linewidth]{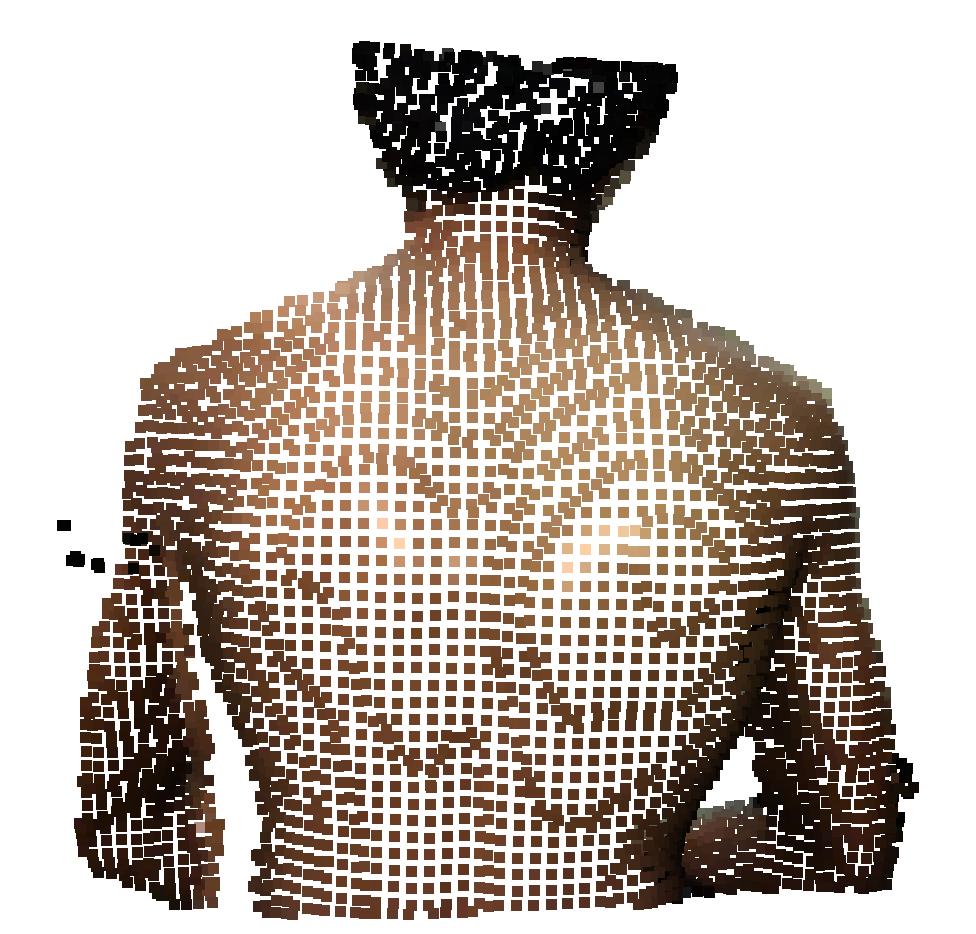}} &
{\includegraphics[width=.325\linewidth]{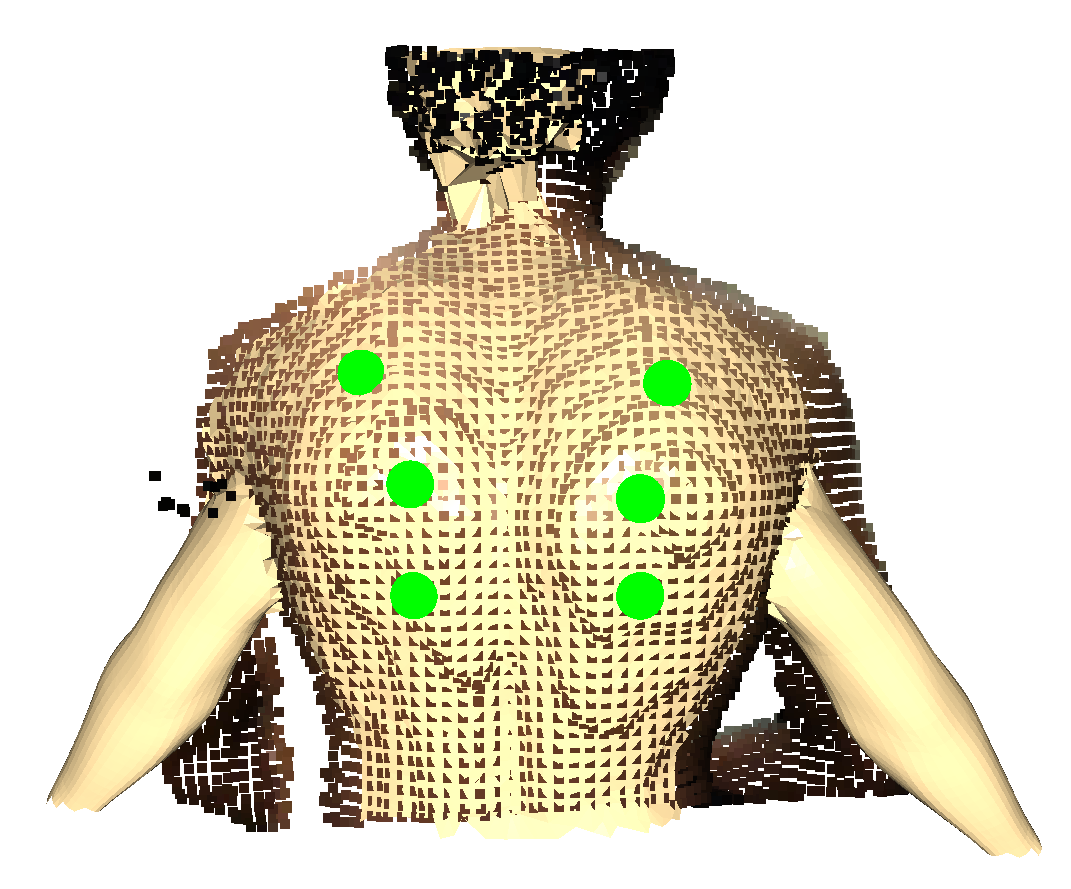}}
\end{tabular}
\vspace{-5px}
\caption{Left to right: labeled reference models, subject (``patient'') scan point clouds, and result of nonrigid registration. Auscultation locations for heart are labeled in red, and lungs in green. \label{fig:registration}}
\vspace{-5px}
\end{figure}

Doctors are trained to auscultate typical points on a patient body as illustrated in~\prettyref{fig:registration}, both for optimal sound quality and ability to diagnose abnormalities on specific structures of the heart and lung. The goal of visual registration is to locate these points on a patient, both to define regions for ausculation and to obtain a prior observation model. \revise{Note that we do not track the patient body after this initial registration is performed and assume that the patient body does not move significantly during auscultation. Small movements are accounted for by the motion controller (Sec.~\ref{sec:motion})}

\subsubsection{Visual Registration}
We first manually label a reference human mesh model with auscultation locations, which we register onto a 3D point cloud of a patient captured with a RGB-D camera. In all of our experiments, we downsample the captured point cloud with a voxel size of 0.01\,m, which strikes a balance between computational time and enough coverage by using a stethoscope whose diameter is approximately  0.02\,m. We first perform rigid registration with RANSAC~\cite{Fischler1981} followed by ICP~\cite{Besl1992} to provide a good initial alignment, then nonrigid registration with nonrigid ICP proposed by Amberg et al.~\cite{Amberg2007}, where a local affine deformation model is used. The initial and final results of one case of visual registration is shown in~\prettyref{fig:registration}. The  regions  of interest on  the  patient  surface  are  limited  to a certain distance $R$ from  the  estimated clinical auscultation  locations~(\prettyref{fig:obs_model}, green circles). 

\subsubsection{Point Cloud Projection and Sound Quality Prior}
\begin{figure}[]
\vspace{-5px}
\centering
\setlength{\tabcolsep}{0px}
\begin{tabular}{cccc}
{\raisebox{0.54\height}{\includegraphics[trim=4cm 2.8cm 3.7cm 3.6cm,clip,width=.32\linewidth]{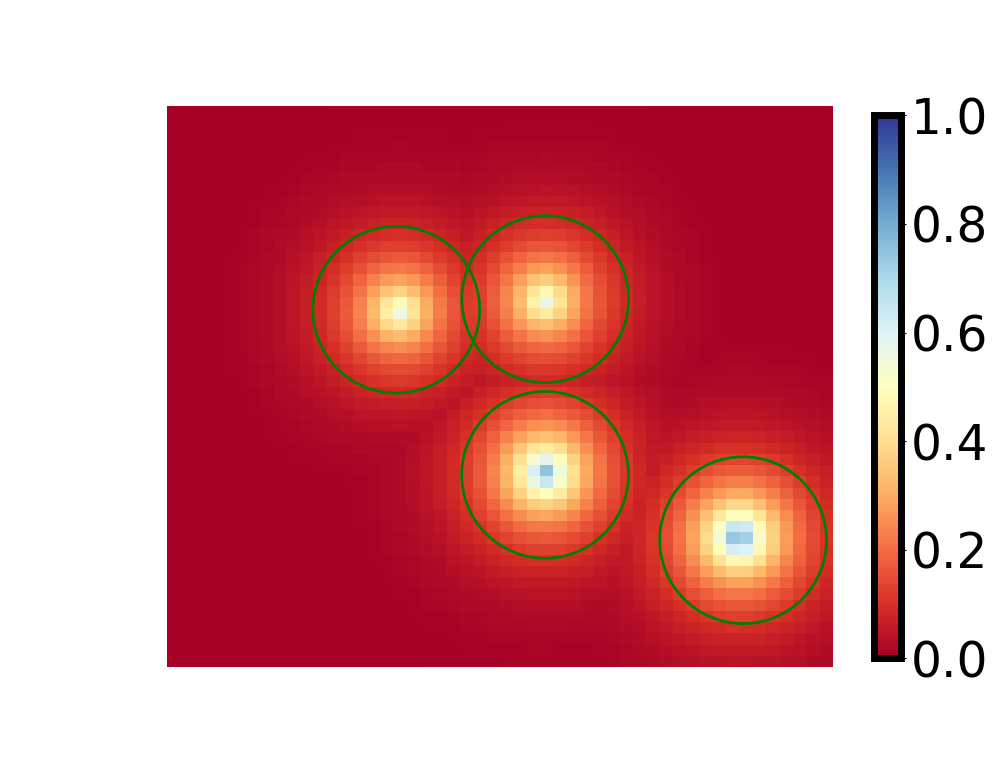}}}\put(-60,93){\scalebox{0.8}{Front Heart}} &
{\raisebox{0.33\height}{\includegraphics[trim=2.7cm 0.5cm 2.5cm 0cm,clip,width=.3\linewidth]{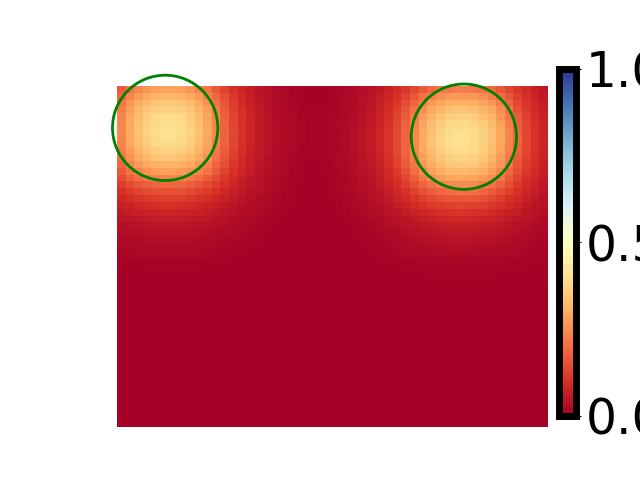}}}\put(-58,93){\scalebox{0.8}{Front Lung}} &
{\raisebox{0.49\height}{\includegraphics[trim=3.0cm 0cm 3.1cm 0cm,clip,width=.26\linewidth]{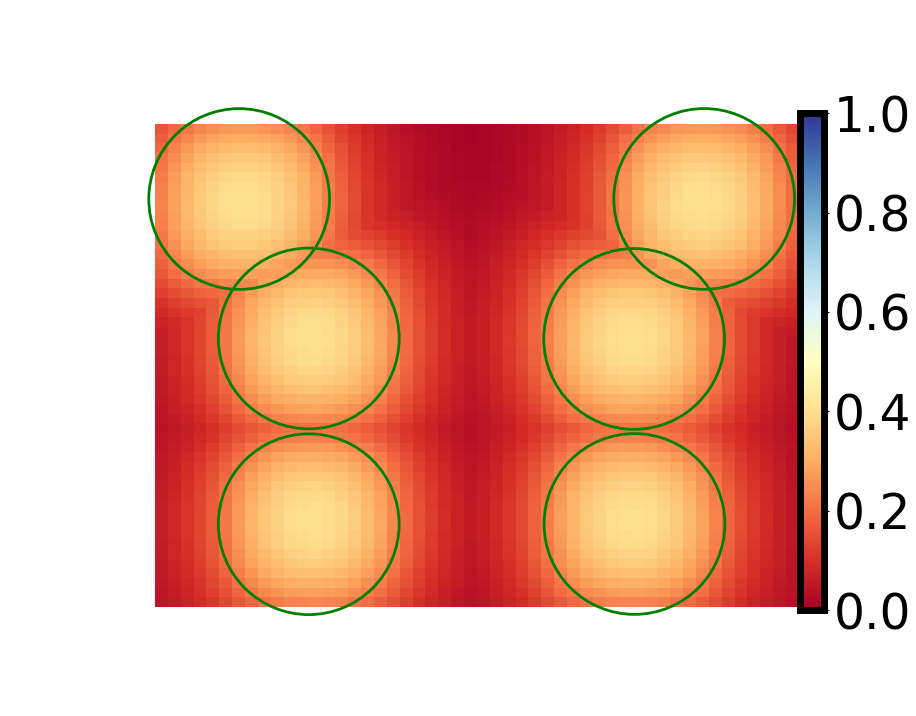}}}\put(-50,93){\scalebox{0.8}{Back Lung}}
\hspace{1px}
{\raisebox{0.13\height}{\includegraphics[trim=20.25cm 0cm 0cm 0cm,clip,width=.066\linewidth]{pics/subject_1_back_lung_prior.png}}}
\end{tabular}
\vspace{-28px}
\caption{Observation model priors for heart and lung on the front of the patient and lung on the back. Each region $A$ is denoted with green circles. [Best viewed in color.]\label{fig:obs_model}}
\vspace{-10px}
\end{figure}

We project the point cloud onto the coronal plane of a patient, and BO only searches amongst locations on this projected plane. The third dimension (height) of the point cloud is associated with each projected point to de-project selected locations on the plane back into 3D space. As discussed in~\prettyref{sec:formulation}, to obtain the prior of the observation model, we first generate a reference quality map $\mu_o(x)$, where we place a negative exponential function at each of the registered auscultation locations in the projected coronal plane. An example of the initial observation model prior is shown in~\prettyref{fig:obs_model}. The latent parameters $\theta = [t_x, t_y, c]$ is initialized to $\theta_o = [0, 0, 1]$, where $t_x$ and $t_y$ are the translation offsets on the projected plane, and $c$ the quality scaling term.

\subsection{Motion Control}\label{sec:motion}
The motion control system generates safe interaction between the stethoscope and the patient, keeping the interaction force at a desired auscultation force $F_{aus}$. According to both Nowak et al.~\cite{Nowak2016} and our observations in experiments, auscultation force $F_{aus}$ has a small effect on sound qualities. We set $F_{aus} = 7N$, which is both comfortable for human subjects and provides good sound qualities in practice. The robot starts at a manually defined home configuration $q_{home}$. For a location $x$ selected by BO, we first get the point $p$ on the patient point cloud where $x$ is projected from. To auscultate at $p$, the robot first moves the stethoscope to an initial position $p_o$ that is a distance $d_o = 0.08$\,m from $p$ in the direction of outward surface normal $\mathbf{n_p}$ at $p$ to account for perception uncertainty, shown in~\prettyref{fig:TRINA}. In particular, we set the target position of the stethoscope frame to be $p_o$, with the $z$-axis aligning with $-\mathbf{n_p}$ and solve for the target robot configuration $q_0$ with inverse kinematics (IK). The robot moves from $q_{home}$ to $q_o$ linearly in the Cartesian space, following maximum linear and angular velocities $v_{max} = 0.065$\,m/s and $\omega_{max} = 1.0$\,rad/s. Starting from $p_o$, the robot moves the stethoscope in the $-\mathbf{n_p}$ direction at constant speed $s = 0.03$\,m/s, during which we also adopt impedance control for enhanced safety, where we utilize the wrist-mounted force-torque sensor on the arm as wrench feedback and control the stethoscope as a mass-spring-damper. The robot keeps moving the stethoscope until the external force in the $\mathbf{n_p}$ direction reaches $F_{aus}$, at which point we record the stethoscope audio for 10\,s and use the learned estimators to estimate the sound quality. After auscultation is finished, the robot first moves back to $q_o$, waits for BO to give the next auscultation location, and moves to the next initial position with a linear Cartesian movement under velocity limits.

\section{Results}
Throughout the experiments, we use the TRINA robot shown in~\prettyref{fig:TRINA}. TRINA is a mobile bimanual manipulator with various visual sensors and swappable end-effectors. In this project, we use a Thinklabs digital stethoscope mounted on the left arm with firm foam as sound insulator. We use the Intel Realsense L515 RGB-D camera on TRINA to capture point clouds. For both the simulation and physical experiments, we use the squared exponential covariance function for GP with length scale $= 0.02$. We chose this length based on our observations of how sound qualities correlate across the surface of the body.

\subsection{Simulation Experiment}
First we compare the performance of different acquisition functions, and evaluate the efficacy of SPAR-GP. To do this, we compare the results of the BO algorithm under visual registration error and patient overall sound quality variations. We first generate a ground truth heart quality map across the patient chest in a similar fashion as the observation model generation in~\prettyref{sec:registration}, with the same shape as~\prettyref{fig:obs_model} but having the negative exponential functions placed at ground truth locations. Then for each setting of acquisition function and prior observation model, we generate a prior by randomly shifting and scaling the ground truth quality map and run the BO algorithm. In particular, the random shift (x,y sampled independently) is uniform between -0.02\,m and 0.02\,m and the random scale is uniform between 0.7 and 1.3. We run this 50 times and compare the average maximum observed qualities across different regions under a given budget $N_{max}$. Simulations are performed for the heart prior only, giving a total of 4 regions. We disable early termination (i.e., Line 6 of Alg.~\ref{alg:BO}) in these experiments.

We set the GP observation noise $\alpha = 0.0417$, and region radius $R = 0.03$\,m. The covariance matrix $\Sigma$ of the prior parameters $\theta$ is a diagonal matrix with entries $\sigma_{t_x}^2 = \sigma_{t_y}^2 = 1.33e{-4}$ (variance of the uniform random distribution of [-0.02,0.02]), and $\sigma_c^2 = 0.03$ (variance of the uniform random distribution of [0.7,1.3]).

Results are summarized in Table I, where we compare 1) acquisition functions EI and UCB with different $\beta$ parameters under different budgets for each of the regions; 2) BO with GP under Zero prior, Residual-GP with Fixed prior, and SPAR-GP (SPAR) prior.  Numbers indicate the maximum sound quality over each region. SPAR-GP consistently outperforms BO with fixed prior and no prior, particularly when the budget of observations is small.  In addition, there is no clear winner among the acquisition functions, so for the subsequent physical experiments we use EI because it is parameter-free. 

\begin{table}[]
\centering
\caption{\label{table:simulation_results} Effects of BO parameters on max sound quality in simulation}
\setlength\tabcolsep{3pt}
\begin{tabular}{@{}p{2cm}llp{1cm}lll@{}}
\toprule
 &
\multicolumn{3}{c}{$N_{max} = 3$} &
\multicolumn{3}{c}{$N_{max} = 10$} \\
 & ~Zero & Fixed & SPAR & ~Zero & Fixed & SPAR \\
\midrule
EI & 0.432 & 0.434 & 0.549 & 0.622 & 0.620 & 0.638\\
UCB, $\beta=0.5$ & 0.370 & 0.428 &  0.550 & 0.629 & 0.624 & 0.660\\
UCB, $\beta=1.0$ & 0.421 & 0.423 & 0.559 & 0.632 & 0.628 & 0.660\\
UCB, $\beta=1.5$ & 0.455 & 0.420 & 0.544 & 0.631 & 0.624 & 0.657 \\
\bottomrule

\end{tabular}
\end{table}

\subsection{Physical Auscultation Experiments}\label{sec:res_estimator}
We compare BO against two other baselines for a complete auscultation session that includes both heart and lung. In RO baseline, the robot automatically auscultate at the registered auscultation landmarks on the patient. In the DT baseline, \revise{a human tele-operates the robot to}  auscultate the patient. We perform each of the three methods once on 4 healthy male human subjects\footnote{This human subjects study was fully reviewed and approved by the University of Illinois at Urbana-Champaign Institutional Review Board, with IRB\#21849.}, whose statistics are listed in~\prettyref{table:bio_stats}.

\begin{table}[]
  \begin{center}
    \caption{\label{table:bio_stats} Statistics of the 4 human subjects}\setlength\tabcolsep{3pt}
    \begin{tabular}{@{}lllll@{}}
    \toprule
       &\textbf{Subject 1} & \textbf{Subject 2} & \textbf{Subject 3} & \textbf{Subject 4}\\
      \midrule
      Weight (kg) & 95.6 &57.4 & 60.5 & 120\\
      Height (m) & 1.84 & 1.82 & 1.81 & 1.83\\
      BMI  & 28.2 &  17.3& 18.5 & 35.8\\
      Age & 33& 22 & 21 & 28\\
    \bottomrule
    \end{tabular}
    \vspace{-10px}
  \end{center}

\end{table}

\subsubsection{Direct Teleoperation (DT)}

In the DT condition, an expert tele-operator performs auscultation at the specified  regions, and adjusts the position of the stethoscope until he/she judges the sound quality to be good enough for making a diagnosis. \revise{The tele-operator is one of the medical school students that labeled the stethoscope recording dataset, and the tele-operator received tele-operation training on TRINA for 3\,hrs prior to the experiments.} Tele-operation on TRINA is achieved via the Oculus Quest VR headset and controllers (Fig.~7(a)), where the VR headset streams the stereo camera on TRINA's head, and the 6D pose of the stethoscope on TRINA is driven in velocity control mode to follow the velocity of one of the VR controllers while a clutch button is depressed.

\begin{figure}[tbp]
\vspace{-5px}
\centering
\setlength{\tabcolsep}{0px}
\begin{tabular}{ccc}
{\raisebox{0\height}{\includegraphics[trim=0cm 0cm 0cm 0cm,clip,width=.26\linewidth]{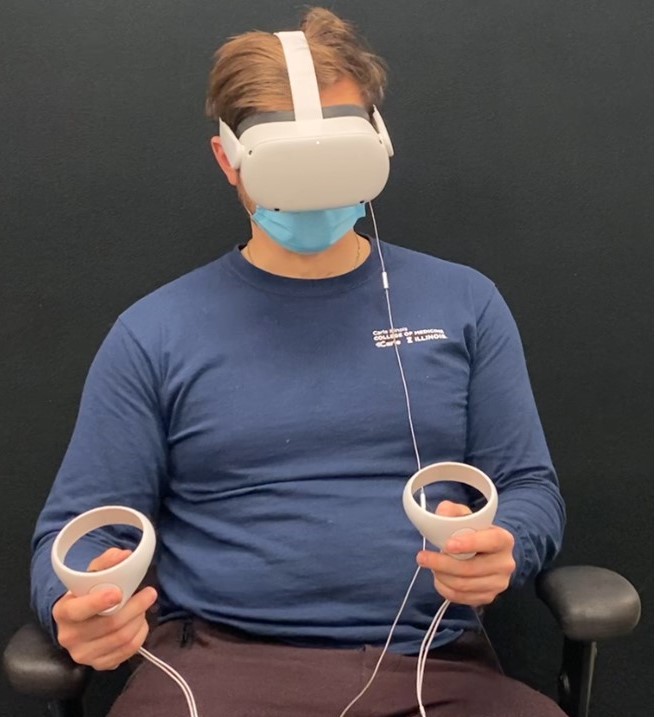}}}\put(-35,-10){\scalebox{0.8}{(a)}} &
{\raisebox{0.\height}{\includegraphics[trim=0cm 0cm 0cm 0cm,clip,width=.36\linewidth]{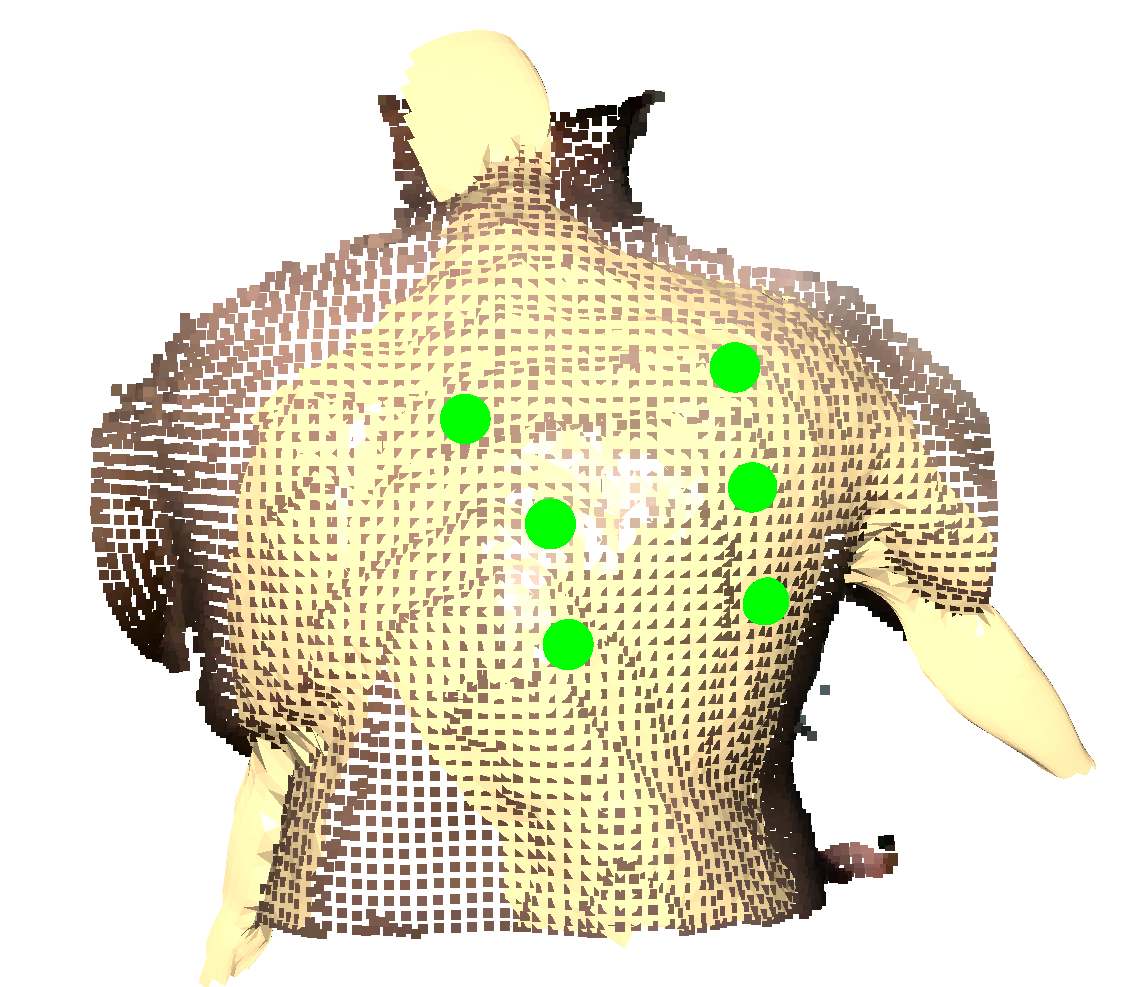}}}\put(-50,-10){\scalebox{0.8}{(b)}} &
{\raisebox{0.05\height}{\includegraphics[trim=0.5cm 0cm 0cm 0cm,clip,width=.33\linewidth]{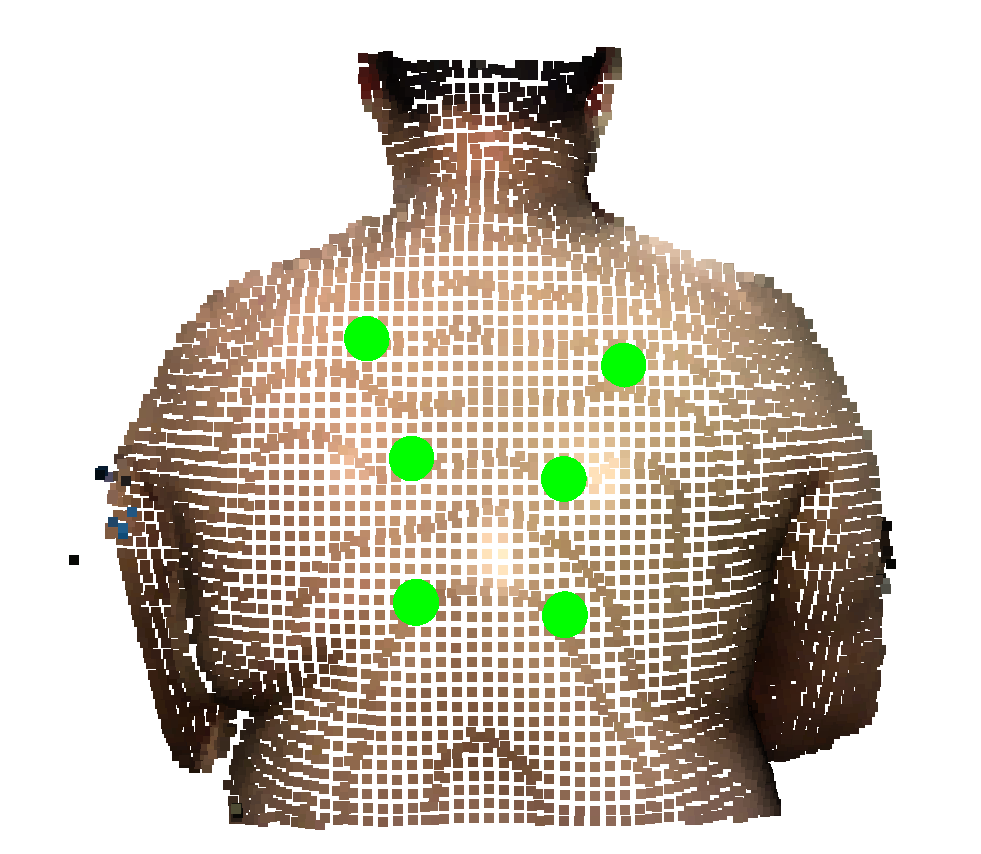}}}\put(-45,-10){\scalebox{0.8}{(c)}}
\end{tabular}

\caption{(a) Tele-operator performing auscultation using VR headset and controllers. (b) Automatic registration failed on the back of subject 4 due to difference in their body type from the reference model. (c) Manual registration on the back of subject 4.}
\vspace{-5px}
\end{figure}

\subsubsection{Bayesian Optimization (BO)}
For BO, we set the acquisition function to EI, $N_{max} = 4$, GP observation noise $\alpha$ = 0.0938 for heart and 0.151 for lung. $\Sigma$ is a diagonal matrix with entries  $\sigma_{t_x}^2 = 2.12e{-4}$, $\sigma_{t_y}^2 = 9.77e{-5}$, and $\sigma_c^2 = 0.03$. In addition, We terminate early if sound quality exceeds $0.5$. This threshold was judged by the 4 raters as the minimum quality for making a confident diagnosis.

The values for $\alpha$ correspond to the maximum inter-rater variance for each recording in the quality estimator dataset.
The values for $R$ and $\Sigma$ were determined empirically as a function of estimated registration error. We estimate  error using a public dataset of complete human scans~\cite{Yang2015}. We select 30 male human mesh models that cover a wide range of body types, with maximum, minimum, and average body mass index (BMI) 34.9, 17.8, and 26.5, respectively. We manually label the nipples, segment the mesh to emulate a RGB-D scan, and apply visual registration. The registration MAE error of the nipples is 0.0173\,m, and the maximum error is 0.0389\,m,  which we set as $R$. 

\subsubsection{Results}
\prettyref{table:results_comparison} breaks down the time spent on visual registration (Reg.), the total amount of time spent during the auscultation session (Total), the total number of auscultations (No.), the average of the maximum rated quality in each region (Avg. Max), and the minimum of the maximum rated quality across all regions (Min Max). Note that sound qualities are average ratings given by the the human raters, not values from the estimator. One example of the final auscultated locations for the heart and lung, and the posterior sound quality estimates are shown in~\prettyref{fig:BO_posterior}.

\begin{table}[ht!]
  \begin{center}
    \caption{\label{table:results_comparison} Experimental results on human subjects}
    \setlength\tabcolsep{3pt}
    \begin{tabular}{@{}llllllll@{}}
      \toprule
      \textbf{Subject} &\textbf{Method} &\textbf{Reg. (s)} & \textbf{Total (s)} & \textbf{No.} &\textbf{Avg. Max}&\textbf{Min Max}\\
      \midrule
     1 &BO & 171 &538 &22 & \textbf{0.727}&\textbf{0.438}\\
      &RO & 171 &370 & 12& 0.648 &0.312\\
      &DT & 0 & 656& 19 & 0.651&0.281\\
      \midrule
     2 &BO & 183 &496 &18 & 0.716 & \textbf{0.531}\\
      &RO & 183 &434 & 12 &0.651 & 0.312\\
      &DT & 0 & 559& 14  & \textbf{0.804} & \textbf{0.531}\\
        \midrule
     3 &BO & 177 &1155 &37 & 0.552 &\textbf{0.375}\\
      &RO & 177 &442 & 12& 0.484 & 0\\
      &DT & 0 & 613& 16 & \textbf{0.576} & 0.25\\
        \midrule
     4$^*$  &BO & 148 &694 &21 & 0.635 & 0.469 \\
      &RO & 148 &495 & 12& 0.599 & 0.375 \\
      &DT & 0 & 468& 14 & \textbf{0.711} & \textbf{0.625}\\
      \bottomrule 
      \multicolumn{7}{l}{\footnotesize *: Manual registration was performed on the back of the subject.}
    \end{tabular}
  \end{center}
\end{table}

\begin{figure}[h]
\vspace{-5mm}
\centering
\setlength{\tabcolsep}{0px}
\begin{tabular}{cccc}
{\raisebox{0.05\height}{\includegraphics[trim=3.5cm 0cm 4cm 3.7cm,clip,width=.325\linewidth]{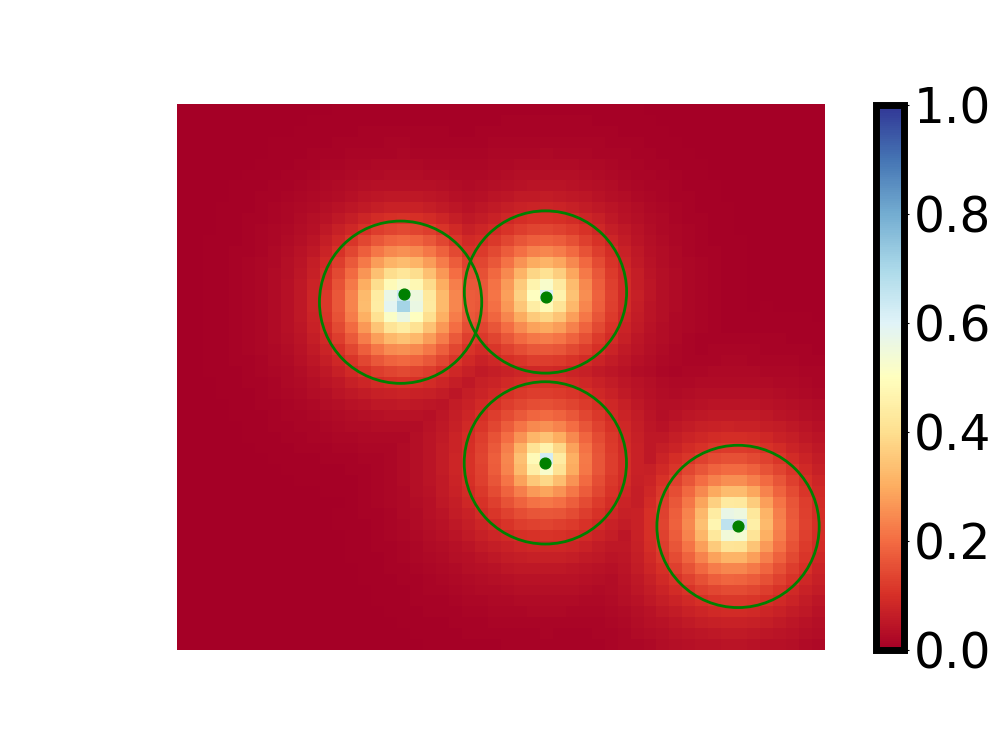}}}\put(-58,77){\scalebox{0.8}{Front Heart}} &
\hspace{-6px}
{\raisebox{0.07\height}{\includegraphics[trim=2.5cm 0cm 2.5cm 0cm,clip,width=.3\linewidth]{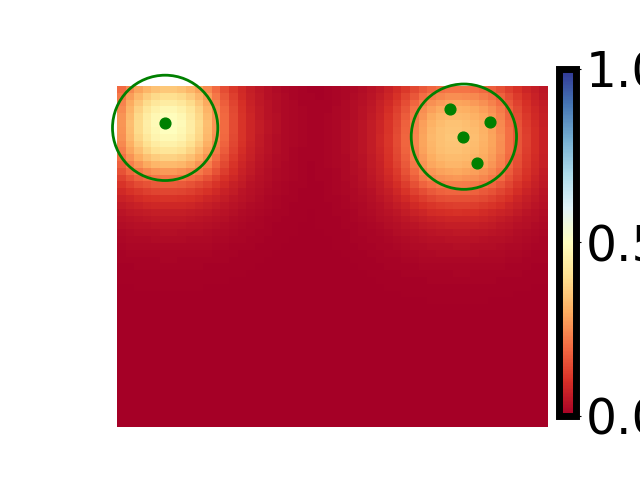}}}\put(-55,77){\scalebox{0.8}{Front Lung}} &
{\raisebox{0.155\height}{\includegraphics[trim=2.5cm 0cm 2.2cm 0cm,clip,width=.28\linewidth]{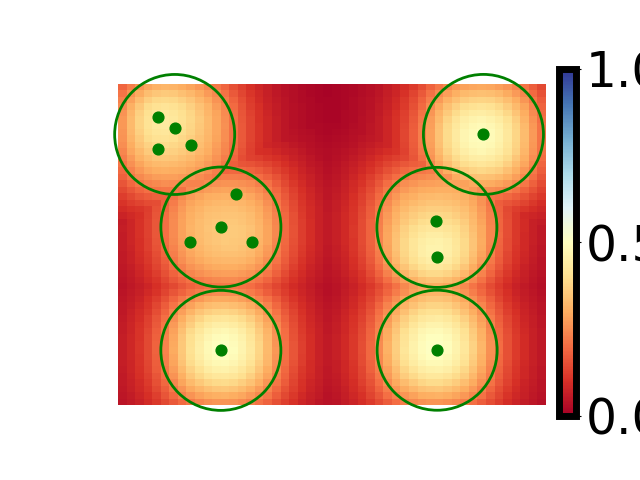}}}\put(-52,77){\scalebox{0.8}{Back Lung}} & 
\hspace{0.2px}
{\raisebox{-0.05\height}{\includegraphics[trim=20.2cm 0cm 0cm 0cm,clip,width=.07\linewidth]{pics/subject_1_back_lung_prior.png}}} 
\end{tabular}
\vspace{-12px}
\caption{Posterior sound quality predictions for heart, front lung, and back lung on Subject 1. Observed locations are marked as green dots.\label{fig:BO_posterior}}
\end{figure}


\revise{We use a linear mixed effects model to evaluate the methods, where we set the best estimated sound quality found in each region to be the dependent variable, methods to be fixed effects, and structure $s$ (heart/lung), auscultation region, and subject to be random effects. The results are presented in~\prettyref{table:fixed_effects}. While the estimated fixed effects coefficients show separations among the methods, they are not strong compared to the confidence intervals. Therefore, we perform a post-hoc pairwise testing using the Tukey's test~\cite{Haynes2013}. The p-values for the pairs BO-DT, BO-RO, and RO-DT are 0.326, 0.0700, and 0.00110, respectively. In addition, all three random effects have a small contribution to the overall variances, where the residual variance dominate.}

\begin{table}[]
  \begin{center}
    \caption{\label{table:fixed_effects} Mixed Effects Model Results ($\beta$:fixed effects coefficient, CI: Confidence Interval, $s$:anatomical structure)}\setlength\tabcolsep{3pt}
    \begin{tabular}{@{}lll|llll@{}}
    \toprule
    \multicolumn{3}{c|}{\underline{\textbf{Fixed Effects}}} &  \multicolumn{4}{c}{\underline{\textbf{Variances of Random Effects}}}\\ 
       \textbf{Method}&\textbf{$\beta$} & \textbf{95\% CI}  & \textbf{$s$} & \textbf{Region} & \textbf{Subject}& \textbf{Residual}\\
      \midrule
      BO & 0.662 & [0.556,0.768]  & \multirow{3}{*}{3.40e-12}& \multirow{3}{*}{1.77e-3}& \multirow{3}{*}{6.03e-3} & \multirow{3}{*}{1.96e-2}\\
      RO & 0.598 & [0.492,0.704]\\
      DT  & 0.703 &  [0.597,0.859]\\
    \bottomrule
    \end{tabular}
  \end{center}
\vspace{-8mm}
\end{table}

\revise{To further understand diagnostic utility, we ask the raters to answer a series of yes-no questions regarding the recordings with the best estimated quality in each region.} \revise{For heart sounds, the key features doctors look for are the first and second heart sound (S1 and S2), while for lung sounds, doctors look for inspiration and expiration. Therefore, for heart, \textit{Q1} and \textit{Q2} are whether the raters are able to hear S1 and S2, respectively. For lung, we ask whether they can hear inspiration (\textit{Q1}) and expiration (\textit{Q2}). Then for both heart and lung recordings, \textit{Q3} asks whether any sounds not originating from the target anatomy could be heard, and \textit{Q4} asks whether continuous noise or static obfuscated the heart/lung sounds. \textit{Q5} asks whether they can make a diagnosis based on the recording. The results are shown in~\prettyref{table:questions}.}

\begin{table}[ht!]
  \begin{center}
    \caption{\label{table:questions} \% of Yes responses for diagnostic utility of recordings Higher is better for Q1, Q2, and Q5 }\setlength\tabcolsep{3pt}
    \begin{tabular}{@{}llllll|lllll@{}}
    \toprule
    \multicolumn{6}{c|}{\underline{\textbf{Heart}}} & \multicolumn{5}{c}{\underline{\textbf{Lung}}} \\ 
    \textbf{Method} &\textit{Q1} & \textit{Q2} & \textit{Q3}& \textit{Q4}& \textit{Q5}&\textit{Q1} & \textit{Q2} & \textit{Q3}& \textit{Q4}& \textit{Q5}\\
    \midrule
    BO & \textbf{100} & \textbf{98.4} &\textbf{12.5} & \textbf{26.6} & \textbf{87.5} &93.0 & 86.7 & 9.38 & \textbf{23.4} & 74.2\\
    RO & 87.5 & 92.2 & 25.0 & 31.3 & 78.1& 89.1 & 74.2 & 11.8 & 24.2 & 66.4\\
    DT & 90.6 & 95.3 & 32.8 & 46.9 & 85.9 & \textbf{96.9} & \textbf{89.8} & \textbf{6.25} & 31.3 & \textbf{85.2}\\
    \bottomrule
    \end{tabular}
  \end{center}
\vspace{-3mm}
\end{table}

\subsection{Discussion}
\revise{Although direct teleoperation (DT) produces better sounds than registration-only automated auscultation (RO) ($p=0.001$), there is no statistically significant difference between DT and Bayesian optimization (BO) ($p=0.326$). BO obtained higher minimum sound quality on patients 1 and 3 (\prettyref{table:results_comparison}), and better diagnostic values for heart sounds (\textit{Q5},~\prettyref{table:questions}). BO was also almost always able to generate S1 and S2 heart sounds (\textit{Q1} and \textit{Q2},~\prettyref{table:questions}). This could indicate greater consistency than a human operator, but more data is needed to test this hypothesis.}
\revise{The evidence suggests that BO achieves higher qualities than RO but is somewhat weak due to the small sample size ($p = 0.070$). BO does achieve a larger minimum quality and better diagnostic values from~\prettyref{table:results_comparison} and~\prettyref{table:questions}. As we expected, adaptive sensing helps overcome uncertainty in visual registration and anatomical differences.}
\revise{Overall diagnostic value (\textit{Q5},~\prettyref{table:questions}) roughly matches the results from the mixed effects model, where DT and BO perform comparable on heart sounds but DT outperforms BO on lung sounds, and both outperform RO. }

Using BO, the robot takes less execution time than DT on subjects 1 and 2, and similar amount on subject 4. However, on subject 3, BO spent much more time auscultating, mainly due to the poor lung qualities observed on the patient.  In fact, the robot ended up using the entire budget $N_{max}$ on all lung locations. Future implementations could skip between regions, and return to past regions if there is sufficient time and need. Moreover, we limited the movement speed of the robot empirically to avoid intimidating the subjects, but further refinement of the motion controller could improve transit times. 

For subject 4, the visual registration for the back exam was identified as having poor alignment. The tele-operator decided to reject the automatic registration, and instead manually labeled the auscultation points during the experiment, as shown in Figs.~7(b) and 7(c). This deviation from fully-automatic procedure is indicated by an asterisk in~\prettyref{table:results_comparison}.  The poor result from the visual registration is likely due to the difference between the body type of the subject (BMI = 35.8) and the reference human model. In future work, visual registration can be improved in various ways, such as adopting multiple reference models for different body types. We also note that raters noticed a moderate amount of noise and static (\textit{Q4},~\prettyref{table:questions}), and we assume this mainly comes from the robot motors because these sounds are not present during manual auscultation. DT was slightly higher than the automated methods, likely due to human tremor. 

Finally, a surprising finding during these studies was that both the tele-operator and the raters detected an incidental cardiac pathology on subject 2. During DT, which were performed before BO and RO, the heart sound was described as a systolic III/IV high pitched crescendo-decrescendo murmur localized to the right-upper sternal border. Upon listening with BO and RO, the murmur was no longer present and the auscultation could be described as a normal S1 and S2 with physiologic S2 splitting distinctly noted, indicating high sound quality in the automated approaches. This murmur's transient nature suggests a few possible diagnoses, including aortic stenosis, bicuspid aortic valve, or a simple, innocent murmur. Subject 2 was recommended to see a physician for further cardiac workup. This is promising since it shows that BO and RO are capable of locating sounds that are of high quality and can be used for diagnosis.

\section{Conclusion and Future Work}
This paper proposed a method that enables a robotic nursing assistant to automatically perform auscultation. To choose sensing locations, we show that using a Bayesian optimization that leverages visual prior information of the clinical auscultation locations outperforms choosing locations according to the prior only, and our system is capable of locating sounds of comparable quality with tele-operation by \revise{a human with clinical auscultation expertise.}

\revise{While the results are promising, our experimental study only includes 4 young human subjects. } In future work a larger-scale study is needed to \revise{evaluate its diagnostic capability on populations likely to exhibit abnormalities}. Furthermore, the system is significantly slower than human doctors, and we intend to accelerate it by making movement faster, and reducing visual registration times, and compare it to manual auscultation. We would also like to improve the robustness of visual registration by using reference human models with varying body types. Finally, we would like to integrate our method with diagnostic algorithms to pre-screen for possible abnormalities.

\printbibliography

@article{vukanovic2006competency,
  title={Competency in cardiac examination skills in medical students, trainees, physicians, and faculty: a multicenter study},
  author={Vukanovic-Criley, Jasminka M and Criley, Stuart and Warde, Carole Marie and Boker, John R and Guevara-Matheus, Lempira and Churchill, Winthrop Hallowell and Nelson, William P and Criley, John Michael},
  journal={Archives of internal medicine},
  volume={166},
  number={6},
  pages={610--616},
  year={2006},
  publisher={American Medical Association}
}

@inproceedings{Driess2019,
author = {Driess, Danny and Hennes, Daniel and Toussaint, Marc},
doi = {10.1109/ICRA.2019.8793773},
isbn = {9781538660263},
issn = {10504729},
booktitle = {IEEE Int'l Conf. Robotics and Automation},
title = {{Active multi-contact continuous tactile exploration with gaussian process differential entropy}},
year = {2019}
}

@article{Ayvali2017,
author = {Ayvali, Elif and Ansari, Alexander and Wang, Long and Simaan, Nabil and Choset, Howie},
doi = {10.1109/LRA.2017.2655619},
issn = {23773766},
journal = {IEEE Robotics and Automation Letters},
number = {2},
pages = {864--871},
publisher = {IEEE},
title = {{Utility-Guided Palpation for Locating Tissue Abnormalities}},
volume = {2},
year = {2017}
}

@article{Grooby2020,
author = {Grooby, Ethan and He, Jinyuan and Kiewsky, Julie and Fattahi, Davood and Zhou, Lindsay and King, Arrabella and Ramanathan, Ashwin and Malhotra, Atul and Dumont, Guy A. and Marzbanrad, Faezeh},
doi = {10.1109/JBHI.2020.3047602},
issn = {21682208},
journal = {IEEE J. Biomedical and Health Informatics},
number = {c},
pages = {1--12},
title = {{Neonatal Heart and Lung Sound Quality Assessment for Robust Heart and Breathing Rate Estimation for telehealth Applications}},
volume = {2194},
year = {2020}
}

@article{Hollinger2013,
author = {Hollinger, Geoffrey A. and Englot, Brendan and Hover, Franz S. and Mitra, Urbashi and Sukhatme, Gaurav S.},
doi = {10.1177/0278364912467485},
issn = {02783649},
journal = {Int. J. Robotics Res.},
number = {1},
pages = {3--18},
title = {{Active planning for underwater inspection and the benefit of adaptivity}},
volume = {32},
year = {2013}
}

@article{Hollinger2013a,
author = {Hollinger, Geoffrey and Sukhatme, Gaurav},
doi = {10.15607/rss.2013.ix.051},
journal = {Robotics: Science and Systems},
number = {5},
title = {{Sampling-based Motion Planning for Robotic Information Gathering}},
volume = {3},
year = {2013}
}

@article{Binney2012,
author = {Binney, Jonathan and Sukhatme, Gaurav S.},
doi = {10.1109/ICRA.2012.6224902},
isbn = {9781467314039},
issn = {10504729},
journal = {IEEE Int'l Conf. on Robotics and Automation},
pages = {2147--2154},
publisher = {IEEE},
title = {{Branch and bound for informative path planning}},
year = {2012}
}

@inproceedings{Amberg2007,
author = {Amberg, Brian and Romdhani, Sami and Vetter, Thomas},
booktitle = {IEEE Conf. Computer Vision and Pattern Recognition},
doi = {10.1109/CVPR.2007.383165},
isbn = {1-4244-1179-3},
issn = {0959-4973},
month = jun,
number = {Supplement 1},
pages = {1--8},
publisher = {IEEE},
title = {{Optimal Step Nonrigid ICP Algorithms for Surface Registration}},
volume = {8},
year = {2007}
}

@article{Salman2018,
archivePrefix = {arXiv},
arxivId = {1711.07063},
author = {Salman, Hadi and Ayvali, Elif and Srivatsan, Rangaprasad Arun and Ma, Yifei and Zevallos, Nicolas and Yasin, Rashid and Wang, Long and Simaan, Nabil and Choset, Howie},
doi = {10.1109/ICRA.2018.8460936},
eprint = {1711.07063},
isbn = {9781538630815},
issn = {10504729},
journal = {IEEE Int. Conf. Robotics and Automation},
pages = {5356--5363},
title = {{Trajectory-Optimized Sensing for Active Search of Tissue Abnormalities in Robotic Surgery}},
year = {2018}
}

@article{Brochu2010,
archivePrefix = {arXiv},
arxivId = {1012.2599},
author = {Brochu, Eric and Cora, Vlad M. and de Freitas, Nando},
journal = {ArXiv},
eprint = {1012.2599},
title = {{A Tutorial on Bayesian Optimization of Expensive Cost Functions, with Application to Active User Modeling and Hierarchical Reinforcement Learning}},
year = {2010}
}

@article{Sainburg2020,
  title={Finding, visualizing, and quantifying latent structure across diverse animal vocal repertoires},
  author={Sainburg, Tim and Thielk, Marvin and Gentner, Timothy Q},
  journal={PLoS computational biology},
  volume={16},
  number={10},
  year={2020},
  publisher={Public Library of Science}
}

@article{Le2020,
  title={Scaling tree-based automated machine learning to biomedical big data with a feature set selector},
  author={Le, Trang T and Fu, Weixuan and Moore, Jason H},
  journal={Bioinformatics},
  volume={36},
  number={1},
  pages={250--256},
  year={2020},
  publisher={Oxford University Press}
}

@inproceedings{Besl1992,
author = {Besl, Paul J. and McKay, Neil D.},
booktitle = {Sensor Fusion IV: Control Paradigms and Data Structures},
doi = {10.1117/12.57955},
editor = {Schenker, Paul S.},
month = apr,
number = {April 1992},
pages = {586--606},
title = {{Method for registration of 3-D shapes}},
volume = {1611},
year = {1992}
}

@article{Fischler1981,
author = {Fischler, Martin A and Bolles, Robert C},
issn = {00010782},
journal = {Graphics and Image Processing},
number = {6},
pages = {381--395},
title = {{Random Sample Paradigm for Model Consensus: A Apphcatlons to Image Fitting with Analysis and Automated Cartography}},
volume = {24},
year = {1981}
}

@article{Giuliani2020,
author = {Giuliani, Manuel and Szcz{\c{e}}{\'{s}}niak-Sta{\'{n}}czyk, Dorota and Mirnig, Nicole and Stollnberger, Gerald and Szyszko, Malgorzata and Sta{\'{n}}czyk, Bartlomiej and Tscheligi, Manfred},
doi = {10.1007/s12553-019-00399-0},
issn = {21907196},
journal = {Health and Technology},
number = {3},
pages = {649--665},
title = {{User-centred design and evaluation of a tele-operated echocardiography robot}},
volume = {10},
year = {2020}
}

@article{Yang2020,
author = {Yang, Geng and Lv, Honghao and Zhang, Zhiyu and Yang, Liu and Deng, Jia and You, Siqi and Du, Juan and Yang, Huayong},
doi = {10.1186/s10033-020-00464-0},
issn = {21928258},
journal = {Chinese J. Mechanical Engineering},
number = {1},
publisher = {Springer Singapore},
title = {{Keep Healthcare Workers Safe: Application of Teleoperated Robot in Isolation Ward for COVID-19 Prevention and Control}},
volume = {33},
year = {2020}
}

@article{Wang2021,
author = {Wang, Xi Vincent and Wang, Lihui},
doi = {10.1016/j.jmsy.2021.02.005},
issn = {0278-6125},
journal = {J. Manufacturing Systems},
number = {February},
publisher = {Elsevier Ltd},
title = {{A literature survey of the robotic technologies during the COVID-19 pandemic}},
year = {2021}
}

@article{Evans2020,
author = {Evans, Kevin D. and Yang, Qian and Liu, Yang and Ye, Ruizhong and Peng, Chengzhong},
doi = {10.1177/8756479320917107},
issn = {15525430},
journal = {J. Diagnostic Medical Sonography},
number = {4},
pages = {370--376},
title = {{Sonography of the Lungs: Diagnosis and Surveillance of Patients With COVID-19}},
volume = {36},
year = {2020}
}

@article{Yu2020,
author = {Yu, R. Z. and Li, Y. Q. and Peng, C. Z. and Ye, R. Z. and He, Q.},
doi = {10.26355/eurrev_202007_22283},
issn = {22840729},
journal = {European Rev. Medical and Pharmacological Sciences},
number = {14},
pages = {7796--7800},
pmid = {32744706},
title = {{Role of 5G-powered remote robotic ultrasound during the COVID-19 outbreak: Insights from two cases}},
volume = {24},
year = {2020}
}

@article{Adams2020,
author = {Adams, Scott J. and Burbridge, Brent and Chatterson, Leslie and McKinney, Veronica and Babyn, Paul and Mendez, Ivar},
doi = {10.1177/1357633X20965422},
issn = {17581109},
journal = {J. Telemedicine and Telecare},
title = {{Telerobotic ultrasound to provide obstetrical ultrasound services remotely during the COVID-19 pandemic}},
year = {2020}
}

@article{Wang2021b,
author = {Wang, Jing and Peng, Chengzhong and Zhao, Yan and Ye, Ruizhong and Hong, Jun and Huang, Haijun and Chen, Legao},
doi = {10.1002/jum.15406},
issn = {15509613},
journal = {J. Ultrasound in Medicine},
pages = {385--390},
pmid = {32725833},
title = {{Application of a Robotic Tele-Echography System for COVID-19 Pneumonia}},
volume = {40},
year = {2021}
}

@article{Mathur2019,
author = {Mathur, Bharat and Topiwala, Anirudh and Schaffer, Saul and Kam, Michael and Saeidi, Hamed and Fleiter, Thorsten and Krieger, Axel},
doi = {10.1109/BIBE.2019.00122},
isbn = {9781728146171},
journal = {IEEE Int. Conf. Bioinformatics and Bioengineering},
pages = {649--656},
title = {{A semi-autonomous robotic system for remote trauma assessment}},
year = {2019}
}

@article{DiLallo2021,
author = {{Di Lallo}, Antonio and Murphy, Robin and Krieger, Axel and Zhu, Junxi and Taylor, Russell H. and Su, Hao},
doi = {10.1109/MRA.2020.3045671},
issn = {1558223X},
journal = {IEEE Robotics and Automation Magazine},
title = {{Medical Robots for Infectious Diseases: Lessons and Challenges from the COVID-19 Pandemic}},
year = {2021}
}

@inproceedings{Nowak2016,
author = {Nowak, Karolina M and Nowak, Lukasz},
booktitle = {Int. Congress on Acoustics},
title = {{On the relation between pressure applied to the chest piece of a stethoscope and parameters of the transmitted bioacoustic signals}},
year = {2016}
}

@article{Yang2015,
author = {Yang, Yipin and Yu, Yao and Zhou, Yu and Du, Sidan and Davis, James and Yang, Ruigang},
doi = {10.1109/3DV.2014.47},
isbn = {9781479970018},
journal = {Int. Conf. 3D Vision},
pages = {41--48},
title = {{Semantic Parametric Reshaping of Human Body Models}},
year = {2015}
}

@article{Koo2016,
author = {Koo, Terry K. and Li, Mae Y.},
doi = {10.1016/j.jcm.2016.02.012},
issn = {15563707},
journal = {J. Chiropractic Medicine},
number = {2},
pages = {155--163},
pmid = {27330520},
publisher = {Elsevier B.V.},
title = {{A Guideline of Selecting and Reporting Intraclass Correlation Coefficients for Reliability Research}},
volume = {15},
year = {2016}
}

@article{Goldman2013,
author = {Goldman, R. E. and Bajo, A. and Simaan, N.},
doi = {10.1017/S0263574712000100},
issn = {0263-5747},
journal = {Robotica},
month = jan,
number = {1},
pages = {71--87},
title = {{Algorithms for autonomous exploration and estimation in compliant environments}},
volume = {31},
year = {2013}
}

@article{arent2017selected,
  title={Selected topics in design and application of a robot for remote medical examination with the use of ultrasonography and ascultation from the perspective of the REMEDI project},
  author={Arent, Krzysztof and Cholewi{\'n}ski, Mateusz and Domski, W and Drwi{\k{e}}ga, M and Jakubiak, Janusz and Janiak, Mariusz and Kreczmer, Bogdan and Kurnicki, Adam and Sta{\'n}czyk, Bart{\l}omiej and Szcz{\k{e}}{\'s}niak-Sta{\'n}czyk, D and others},
  journal={J. Automation Mobile Robotics and Intel. Sys.},
  volume={11},
  number={2},
  pages={82--94},
  year={2017}
}

@inproceedings{LiTRINASystem2017,
	author = {Zhi Li and Peter Moran and Qingyuan Dong and Ryan J Shaw and Kris Hauser},
	title = {Development of a Tele-Nursing Mobile Manipulator for Remote Care-giving in Quarantine Areas},
	booktitle = {IEEE Int. Conf. Robotics and Automation},
	year = {2017}
}

@article{Kim2021,
author = {Kim, Ria and Schloen, John and Campbell, Nathan and Horton, Samantha and Zderic, Vesna and Efimov, Igor and Lee, David and Park, Chung Hyuk},
doi = {10.1109/TMRB.2020.3047154},
issn = {25763202},
journal = {IEEE T. Medical Robotics and Bionics},
number = {1},
pages = {96--105},
title = {{Robot-Assisted Semi-Autonomous Ultrasound Imaging with Tactile Sensing and Convolutional Neural-Networks}},
volume = {3},
year = {2021}
}

@article{Khamis2021,
author = {Khamis, Alaa and Meng, Jun and Wang, Jin and Azar, Ahmad Taher and Prestes, Edson and Li, Howard and Hameed, Ibrahim A. and Tak{\'{a}}cs, {\'{A}}rp{\'{a}}d and Rudas, Imre J. and Haidegger, Tam{\'{a}}s},
doi = {10.12700/APH.18.5.2021.5.3},
issn = {17858860},
journal = {Acta Polytechnica Hungarica},
number = {5},
pages = {13--35},
title = {{Robotics and intelligent systems against a pandemic}},
volume = {18},
year = {2021}
}

@article{Haidegger2019,
author = {Haidegger, Tamas},
doi = {10.1109/tmrb.2019.2913282},
journal = {IEEE T. Medical Robotics and Bionics},
number = {2},
pages = {65--76},
title = {{Autonomy for Surgical Robots: Concepts and Paradigms}},
volume = {1},
year = {2019}
}

@article{Wu2012,
author = {Wu, Tingfan and Movellan, Javier},
doi = {10.1109/IROS.2012.6385977},
isbn = {9781467317375},
issn = {21530858},
journal = {IEEE Int‘l Conf. on Intelligent Robots and Systems},
pages = {725--731},
title = {{Semi-parametric Gaussian process for robot system identification}},
year = {2012}
}

@article{Camoriano2016,
archivePrefix = {arXiv},
arxivId = {1601.04549},
author = {Camoriano, Raffaello and Traversaro, Silvio and Rosasco, Lorenzo and Metta, Giorgio and Nori, Francesco},
doi = {10.1109/ICRA.2016.7487177},
eprint = {1601.04549},
isbn = {9781467380263},
issn = {10504729},
journal = {IEEE Int. Conf. Robotics and Automation},
pages = {544--550},
title = {{Incremental semiparametric inverse dynamics learning}},
volume = {2016-June},
year = {2016}
}

@article{Ko2007,
author = {Ko, Jonathan and Klein, Daniel J. and Fox, Dieter and Haehnel, Dirk},
doi = {10.1109/ROBOT.2007.363075},
isbn = {1424406021},
issn = {10504729},
journal = {IEEE Int'l Conf. on Robotics and Automation},
number = {April},
pages = {742--747},
title = {{Gaussian processes and reinforcement learning for identification and control of an autonomous blimp}},
year = {2007}
}

@Inbook{Haynes2013,
author="Haynes, Winston",
editor="Dubitzky, Werner
and Wolkenhauer, Olaf
and Cho, Kwang-Hyun
and Yokota, Hiroki",
title="Tukey's Test",
bookTitle="Encyclopedia of Systems Biology",
year="2013",
publisher="Springer New York",
address="New York, NY",
pages="2303--2304",
isbn="978-1-4419-9863-7",
doi="10.1007/978-1-4419-9863-7_1212"
}
\end{document}